
\let\dvips=\relax        

\catcode`\^^c2=13
\ifx\files\undefined \input index \fi

\catcode`\@=11
\let\utf=\relax
\let\utf@Ch=\relax
\let\stringate=\relax
\let\dohigh=\relax
\let\doaccents=\relax
\let\dosymbols=\relax
\let\deactivate=\relax
\def\stringaccents{\def\'{\string\'}\def\~{\string\~}%
 \def\"{\string\"}\def\`{\string\`}\def\^{\string\^}}
\catcode`\@=12

\files


\ifx\loadfont\undefined \input fonts \fi
\xiifonts \xiititles \rm
\font\sf=cmss12
\font\sfit=cmssi12

\catcode`\@=11

\font\xiibb=msbm10 at12pt
\font\ixbb=msbm9
\font\vibb=msbm6
\newfam\bbfam 
\textfont\bbfam=\xiibb \scriptfont\bbfam=\ixbb
\scriptscriptfont\bbfam=\vibb

\mathchardef\subsetneq="3828

\font\xiifrak=eufm10 at12pt
\font\ixfrak=eufm9
\newfam\frakfam \textfont\frakfam=\xiifrak \scriptfont\frakfam=\ixfrak

\mathcode`?="603F 
\def\ifmath$#1${\relax\ifmmode #1\else$#1$\fi}


\ifcase\pdfoutput
 
\else
 \pdfcompresslevel=9
 \pdfdecimaldigits=4
 \pdfhorigin=1truein
 \pdfvorigin=1truein
 \pdfpageheight=297truemm
 \pdfpagewidth=210truemm
\fi
\hsize=15.92truecm \vsize=24.62truecm 
\advance\vsize -30pt

\def\twodigits#1{\ifnum #1<10 0\fi \number#1}

\let\version=\todayiso
\def\Folio{\ifnum\pageno<0
 \uppercase\expandafter{\romannumeral-\pageno}\else\number\pageno\fi}

\headline={\vrule height10.2pt depth4.2pt width0pt
 {\xiitt www.ramoncasares.com}\quad{\xiirm\version}\hfil
 \quad{\xiibf\jobname\quad\Folio}}%
\def\makeheadline{\vbox to 30pt{\colorblack\line{\the\headline}%
  \kern 1pt \hrule height 1pt\vfil\endcolor}\nointerlineskip}
\nopagenumbers


\newcount\secno
\newcount\ssecno
\newcount\thno
\newcount\parno
\let\presec=\empty
\def\presec{\S} 

\parskip=0pt plus 1pt
\newdimen\oldparindent \oldparindent=20pt \parindent=0pt
\def\hang{\hangindent\oldparindent}

\def\numberedpars{\global\advance\parno1 
 \noindent\hbox to\oldparindent{{\xiiscriptsy\char123
  \xiiscriptrm\number\parno}\hfil$\cdot$\hfil}\ignorespaces}

\def\begincenter{\par\begingroup \parindent=0pt
 \advance\leftskip 0pt plus 1fill \advance\rightskip 0pt plus 1fill
 \def\\{\unskip\break\ignorespaces}}
\def\endcenter{\par\endgroup}

\outer\def\section#1{\vskip0pt plus 90pt\penalty-500\vskip0pt plus-90pt
 \everypar{}\advance\secno1
 \ssecno=0 \thno=0 \parno=0
 \goodbreak\vskip 2pc plus 1pc
 \def\secid{\presec\the\secno}
 \noindent{\fonttwo\secid\ #1\dest\toc1{#1}\lbl{#1}{\secid}}\par
 \everypar{\numberedpars}}
\outer\def\xsection#1{\vskip0pt plus 90pt\penalty-500\vskip0pt plus-90pt
 \everypar{}\parno=0
 \goodbreak\vskip 2pc plus 1pc
 \def\secid{}
 \noindent{\fonttwo#1\dest\toc1{#1}\lbl{#1}{}}\par
 \everypar{\numberedpars}}


\outer\def\subsection#1{\vskip0pt plus 60pt\penalty-250\vskip0pt plus-60pt
 \everypar{}\advance\ssecno1 \par
 \ifnum\parno=0 \vskip 0.5pc plus 3pt minus 3pt\else\vskip 1pc plus 6pt minus 6pt\fi
 \thno=0 \parno=0
 \def\secid{\presec\the\secno.\the\ssecno}%
 \noindent{\fontthree\secid\ #1\dest\toc2{#1}\lbl{#1}{\secid}}\par
 \everypar{\numberedpars}}

\outer\def\clause#1{\vskip0pt plus 30pt\penalty-150\vskip0pt plus-30pt
 \everypar{}\medskip\hang\noindent
 \advance\thno1 \def\secid{\presec\the\secno.\the\ssecno.\the\thno}%
 {\sc\secid\ #1}\quad\ignorespaces}

\def\note#1{\ifvmode
 \vtop to0pt{\vss\rlap{\color{White}\xiiscriptrm #1\endcolor}\kern-2pt\null}%
\else
 \vadjust{\vtop to0pt{\vss
 \rlap{\color{White}\xiiscriptrm #1\endcolor}\kern-2pt\null}}\fi}
\def\label#1{\note{#1}\dest\lbl{#1}{\secid}\ignorespaces}


\def\title#1{\def\titledoc{#1}}
\def\author#1{\def\authordoc{#1}}
\def\contact#1{\def\contactdoc{#1}}
\def\keywords#1{\def\keywordsdoc{#1}}
\def\subject#1{\def\subjectdoc{#1}}

\def\beginnote{\insert\footins\bgroup\strut\colorblack}
\def\endnote{\endcolor\egroup}

\def\maketitle{\par
 \begingroup
 \def\'##1{\if##1o\string^^f3\else ##1\fi}
 \def\\{ }%
 \ifcase\pdfoutput
  \ifx\dvips\undefined
  \else
   \special{ps: [
    /Title (\titledoc) /Author (\authordoc)
    /Keywords (\keywordsdoc) /Subject (\subjectdoc)
    /DOCINFO pdfmark }%
  \fi
 \else
  \pdfinfo{/Title (\titledoc) /Author (\authordoc)
    /Keywords (\keywordsdoc) /Subject (\subjectdoc)}\fi
 \endgroup
 \null\vskip 3pc plus 4pc minus 1pc
 \def\secid{}\dest\toc0{\titledoc}\lbl{\titledoc}{}
 \begincenter \fontzero\titledoc \endcenter
 \vskip 1pc plus 1pc
 \centerline{\fontone\authordoc}
 \ifx\contactdoc\undefined\else
  \vskip 6pt minus 3pt
  \centerline{\rm\contactdoc}\fi
 \vskip 2pc plus 1pc minus 1pc\relax}

\def\beginabstract{\begingroup \parindent=55pt\narrower \sl\parindent=20pt\noindent}
\def\endabstract{\par \ifx\keywordsdoc\undefined\else \sfit\rightskip=55pt plus3em
 \smallskip\setbox0=\hbox{Keywords:\quad}\hangindent\wd0
 \noindent\box0\keywordsdoc\par\fi \smallskip\endgroup}



\def\citenone #1 (#2){\ref{#1#2}}
\def\citetext [#1] #2 (#3){\refx{{\def\1{#2}\def\2{#3}#1}}{#2#3}}
\def\citestar * #1 (#2){\refx{#1\ #2}{#1#2}}

\def\cite{\futurelet\nexttoken\citexx}
\def\citexx{\if *\nexttoken \let\next=\citestar \else
 \if [\nexttoken \let\next=\citetext \else
 \let\next=\citenone \fi\fi\next}

\def\chreference #1 (#2){\everypar{}\par
 \vskip0pt plus 2\baselineskip\penalty-43
 \vskip0pt plus-2\baselineskip
 \noindent\hangindent20pt\relax
 #1\thinspace(#2)\dest\lbl{#1#2}{#1\ (#2)}}

\def\brreference #1 (#2){\everypar{}\par
 \vskip0pt plus 2\baselineskip\penalty-43
 \vskip0pt plus-2\baselineskip
 \noindent\hangindent20pt\relax
 [#1 #2]\dest\lbl{#1#2}{[#1 #2]}}

\let\reference=\chreference

\def\references{\vskip0pt plus 90pt\penalty-500\vskip0pt plus-90pt
 \everypar{}\parno=0
 \goodbreak\vskip 2pc plus 1pc
 \def\secid{}
 \noindent{\fonttwo References\dest\toc1{References}\lbl{References}{}}\par
 \begingroup
  \def\reference ##1 ({{\stringaccents\def\&{\string\&}\xdef\1{##1}}\next}
  \def\next ##1) ##2\par{
    \expandafter\gdef\csname \1##1\endcsname{##2}}
  \input RCbiblio
 \let\item=\bibitem \let\xitem=\xbibitem
 \everypar{}\smallskip \parskip=0pt \frenchspacing}

\def\bibitem #1 (#2){\chreference #1 (#2):\enskip
 {\stringaccents\def\&{\string\&}\xdef\1{#1}}\csname \1#2\endcsname}
\def\xbibitem #1 (#2) = #3 (#4){\chreference #1 (#2):\enskip
 {\stringaccents\def\&{\string\&}\xdef\1{#3}}\csname \1#4\endcsname}

\def\book#1{{\em#1}}
\def\periodical#1{{\em#1}}
\def\ISBN{{\sc isbn: }}
\def\DOI#1{{\sc doi: }\URL{#1}<https://dx.doi.org/#1>}
\def\DOIx#1#2{{\sc doi: }\URL{#1}<#2>}


\def\tocline#1#2#3#4#5{\par \ifcase#1
 \toclinezero{#2}{#3}{#4}{#5}\or
 \toclineone{#2}{#3}{#4}{#5}\or
 \toclinetwo{#2}{#3}{#4}{#5}\else
 \toclinethree{#2}{#3}{#4}{#5}\fi}

\def\gobblefour#1#2#3#4{\relax}
\def\leaderfill{\leaders\hbox to\baselineskip{\hss.\hss}\hfill}
\def\toclinezero#1#2#3#4{\centerline{\def\\{ }\goto{\fonttwo #1}{#2}}\bigskip}
\def\toclineone#1#2#3#4{\def\3{#3}%
 \line{\fontthree \ifx\3\empty\else #3 \fi#1\leaderfill \goto{#4}{#2}}}
\def\toclinetwo#1#2#3#4{\def\3{#3}%
 \line{\qquad \ifx\3\empty\else #3 \fi#1\leaderfill \goto{#4}{#2}}}
\let\toclinethree=\gobblefour

\newread\tocfile


\def\strutdepth{\dp\strutbox}
\def\marginalstar{\strut\vadjust{\kern-\strutdepth\specialstar}}
\let\marginalsymbol=*
\def\specialstar{\vtop to \strutdepth{
 \baselineskip\strutdepth
 \vss\llap{\rm\truecolor{Red}\marginalsymbol\endcolor\quad}\null}}

\def\new/{\truecolor{Red}}
\def\wen/{\endcolor\ifhmode\marginalstar\fi}

\def\uncatcodespecials{\def\do##1{\catcode`##1=12 }\dospecials}
\def\del/{\begingroup\uncatcodespecials
 \ifvmode \let\next=\DELv \else \let\next=\DELh \fi\next}
{\catcode`\|=0 \catcode`\\=12
 |long|gdef|DELv#1\led/{|endgroup}
 |long|gdef|DELh#1\led/{|marginalstar|endgroup|ignorespaces}}
\def\led/{\errmessage{Error! Unnested led.}}



\def\uncatcodespecials{\def\do##1{\catcode`##1=12 }\dospecials}

\def\verb{\begingroup\setupverbatim\d@verbatim}
\def\setupverbatim{\tt\parskip0pt plus0.1pt minus0.1pt
 \everypar={}\def\par{\leavevmode\endgraf}\catcode`\`=13
 \uccode`\~=`\ \uppercase{\let~=\ }\uccode`\~=`\`
 \uppercase{\def~{\relax\lq}}\uccode`\~=0
 \obeylines\uncatcodespecials\obeyspaces \verbatimoptions}
\def\d@verbatim#1{\def\next##1#1{##1\endgroup}\next}
\def\verbatimoptions{}
{\catcode`\`=13 \gdef`{\lq}} 


\def\newline{\ifvmode\null\else\null\hfil\break\fi}
\let\\=\newline

\def\begincitation{\par\begingroup\everypar={}\sl
 \advance\leftskip20pt\advance\rightskip20pt}

\def\needspace#1{\vskip0pt plus #1\penalty-250\vskip0pt plus -#1\relax}

\def\em{\ifdim \fontdimen1\font=0pt \aftergroup\smartitc@r \fi \it}
\def\qt{\ifdim \fontdimen1\font=0pt \aftergroup\smartitc@r \fi \sl}
\def\smartitc@r{\ifhmode \expandafter\itpuncl@ok \fi}
\def\itpuncl@ok{\begingroup\futurelet\ITCt@mpa\itcort@st}
\def\itcort@st{\def\ITCt@mpb{\ITCt@mpa}%
 \ifcat\noexpand\ITCt@mpa,\setbox0=\hbox{\ITCt@mpb}%
  \ifdim\ht0<0.3ex \let\itc@rdo=\endgroup \fi\fi \itc@rdo}
\def\itc@rdo{\skip0=\lastskip \ifdim\skip0=0pt \/\else
 \unskip \/\hskip\skip0 \fi \endgroup}

\def\definition#1{{\em #1}}

\def\URL{\leavevmode\begingroup\catcode`\#=12\catcode`\_=12\relax\@RL}
\def\@RL#1<#2>{\def\1{#1}\ifx\1\empty\def\2{#2}\else\def\2{#1}\fi
\ifcase\pdfoutput
 \ifx\dvips\undefined
  {\color{Red}\2\endcolor}\else
  {\color{Red}\2\endcolor}%
\fi
\else
 \pdfstartlink attr{/Border [0 0 0]}
  user{/Subtype /Link /A << /Type /Action
  /S /URI /URI (#2) >>}{\color{Red}\2\endcolor}\pdfendlink
\fi\endgroup\relax}

\def\MTfigures{\ifx\MT\undefined \input metatex \fi
 \MTline{if known prologues:} 
 \MTline{ verbatimtex}
 \MTline{ \\input fonts \\xiifonts}
 \MTline{ \\catcode`\\@=11} 
 \MTline{ etex}
 \MTline{fi}
 \MTline{}}

\catcode`\@=12

\def\version{20190630}   

\let\plainend=\end
\def\bye{\par\vfill\supereject\plainend}
\def\begin#1{\csname begin#1\endcsname\ignorespaces}
\def\end#1{\csname end#1\endcsname\ignorespaces}
\let\plainitem=\item
\def\circleditem{\plainitem{$\circ$}}
\def\beginitemize{\begingroup
 \let\item=\circleditem
 \everypar{}\parindent=20pt\parskip=0pt plus 1pt\relax}

\darkcolors              
\def\note#1{\relax}      

\def\sep{\leavevmode\hbox{\tt\char`\ }}

\hyphenation{pro-p-o-si-tions Ent-schei-dungs-prob-lem}


\title{Syntax Evolution: Problems and Recursion}
\author{Ram\'on Casares}
\contact{{\sc orcid:}
\URL 0000-0003-4973-3128<http://orcid.org/0000-0003-4973-3128>}
\subject{Linguistics, Computing, Cognition.}
\keywords{syntax evolution,
          problem solving,
          Turing completeness,
          recursion,
          language and thinking.}

\beginnote 
This is
\DOI{10.6084/m9.figshare.4956359}.v2,
version {\tt\version}.\\
\copyright\ 2017, 2019 Ram\'on Casares; licensed as {\tt cc-by}.\\
Any comments on it to
{\tt\URL papa@ramoncasares.com<mailto:papa@ramoncasares.com>}
are welcome.
\endnote

\maketitle


\beginabstract
To investigate the evolution of syntax,
we need to ascertain the evolutionary r\^ole of syntax
and, before that, the very nature of syntax.
Here, we will assume that syntax is computing.
And then, since we are computationally Turing complete,
we meet an evolutionary anomaly, the anomaly of syntax:
we are syntactically too competent for syntax.
Assuming that problem solving is computing, and
realizing that the evolutionary advantage of Turing completeness
is full problem solving and not syntactic proficiency,
we explain the anomaly of syntax by postulating that
syntax and problem solving co-evolved in humans
towards Turing completeness.
Examining the requirements that full problem solving impose
on language, we find firstly that
semantics is not sufficient and that
syntax is necessary to represent problems.
Our final conclusion is that
full problem solving requires a functional semantics
on an infinite tree-structured syntax.
Besides these results,
the introduction of Turing completeness and problem solving
to explain the evolution of syntax should help us to fit
the evolution of language within the evolution of cognition,
giving us some new clues to understand the elusive relation
between language and thinking.
\endabstract

\vfill\break
\null\vfill
\openin\tocfile=auxiliar.toc
\ifeof\tocfile
 \closein\tocfile
 \message{No toc file!}%
\else
 \closein\tocfile
 \input auxiliar.toc
\fi
\vfill\break

\section{Introduction}

\subsection{The Anomaly of Syntax}

Why did only we humans evolve Turing completeness?
That is the question which drives my investigations.
It is a technical question that assumes
that the members of our own species {\it Homo sapiens\/} are Turing complete,
and that only our species is Turing complete,
and then it demands some previous clarifications
about Turing completeness and about the assumptions.

Turing completeness is the maximum computing power,
and computing is a syntactic activity,
because computing consists in transforming
strings of symbols following syntactic rules.
The rules are syntactic because
they do not take into account the symbols meanings,
nor the computer intentions,
but only the symbols themselves and
their positions in the strings.
So we will start in Section \ref{Syntax} with syntax,
taking the main ideas in \cite Chomsky (1959),
namely that syntax is computing and
that then we should
locate natural language syntax in a computing hierarchy.
Natural language syntax seems to be mildly context sensitive,
but for us here it is enough to acknowledge that it is decidable,
as it should be to guarantee that its
generation and parsing are free of infinite loops.
Summary of \ref{Syntax}:
syntax is computing, and
natural language syntax is decidable.

Computing was founded by \cite Turing (1936)
to serve as a mathematical model of problem solving.
The model is successful because it abstracts away
the limitations of a device in memory and speed
from its computing capacity,
and because we humans exhibit that
computing capacity completely.
So in Section \ref{Problem Solving}
we will see that
problem solving is computing, that
Turing completeness is the maximum computing power,
and then it is the maximum problem solving power, and that
we humans are Turing complete
because we can calculate whatever any
Turing machine can compute.

Computing is just one version of recursion,
so in Section \ref{Recursion} we will
deal with recursive devices, languages and functions.
For us here, only Turing complete devices are recursive, and
only the languages used to program Turing complete devices
are recursive, while any computable function is recursive.
A mathematical fact is that
every Turing complete device is undecidable,
but we show that the syntax of a recursive language
can be decidable. We also show that,
to compute any recursive function,
the recursive language of the recursive device needs
a functional semantics, which is a semantics
that can provide the meaning of any recursive function
without exceptions.
And the characteristic property of Turing complete devices,
its full programmability, indicates us that
we are the only Turing complete species,
because only we can learn any natural or artificial language,
as English or {\sc Lisp}.

At this point an evolutionary anomaly,
which we will call the anomaly of syntax,
should be apparent, because it seems that
we are computationally, which is to say syntactically,
 too competent for syntax:
Why would evolution select and keep
a capacity to generate and parse {\it any\/} language,
when generating and parsing {\it one\/} language would
be enough to survive?
Why would evolution select and keep
our {\it undecidable\/} Turing complete syntactic capability
to perform a much simpler {\it decidable\/} syntax?

If syntax does not require Turing completeness,
then some other activity should require it.
So let us take any possible activity, say vision.
But our vision is not better than bonobos vision, and
it is less demanding than eagles vision, and therefore
vision cannot explain why only we evolved Turing completeness.
Then, the activity requiring Turing completeness
has to be something that no other species is doing,
and speaking a language with syntax is something
that no other species is doing,
but then we have to face the anomaly of syntax,
because syntax does not require Turing completeness.

The anomaly of syntax is a narrow formulation
of  Wallace's problem; in his words,
``Natural selection could only have endowed the savage
with a brain a little superior to that of an ape
whereas he possesses one very little inferior
to that of an average member of our learned societies'',
which we requote from \cite Bickerton (2014), page~1.
Most people disregard Wallace's problem and the
anomaly of syntax because otherwise a complex
capacity as recursion, also known as Turing completeness,
could not have an evolutionary cause.
Only the most consequent scientists, as Chomsky,
would follow the anomaly implications until its final
consequences. As understanding Chomsky's reasons
always sheds light on deep problems, we will try
to explain his position on the anomaly of syntax.

In our case, we avoid the anomaly of syntax
using a coincidence already suggested:
syntax is computing, and 
problem solving is computing, too.
Then our computing machinery would have evolved
to satisfy requirements coming both 
from syntax and from problem solving;
in other words,
syntax and problem solving co-evolved in humans.
And though syntax does not require Turing completeness,
full problem solving does.
Now, we can answer part of our original question:
humans evolved Turing completeness
to solve as many problems as possible.

\subsection{The Evolution of Syntax}

If solving as many problems as possible explains
why we evolved Turing completeness,
and it is so good,
why did {\it only\/} we humans evolve Turing completeness?
Our answer here to this why-{\it only\/}-we question
will be more tentative and less informative than
the answer to the why-we question,
and we will address it
by proposing how could have been
our evolutionary road to recursion 
under the hypothesis that
syntax and problem solving co-evolved in humans
towards Turing completeness.
So we will start to answer the why-{\it only\/}-we question
in Section \ref{Evolution}
by examining the requirements
that full problem solving impose on language.

Firstly, we will learn that
to express problems we need variables,
which are words free of meaning,
and sentences, because separate words are not enough.
Therefore, semantics is not sufficient and
syntax is necessary to represent problems.

Then, we will examine the two kinds of conditions
that full problem solving impose on language:
those relating to data structures, that require
an infinite syntax able to deal with hierarchical tree structures,
and those other relating to computing capacity,
that require Turing completeness.
Therefore, our conclusion will be that
full problem solving requires a functional semantics
on an infinite tree-structured syntax.
This way, the full problem solver can represent any problem,
and it can calculate any computable function without restrictions,
so it can calculate any way of solving problems.
A {\sc Lisp} interpreter will serve us to show how,
just by meeting the requirements of full problem solving,
the result is a recursive device.

In the next section, \ref{Discussion},
we will discuss the meaning of all that we have seen,
so we can try to answer our question.
As syntax uses hierarchical tree structures of words,
then words are a prerequisite for syntax, and therefore
we will assume that
the starting point of the evolution of syntax
towards Turing completeness was a protolanguage
with words but without sentences.
$$\hbox{Animal Communication}\longrightarrow
 \hbox{Protolanguage}\longrightarrow
 \hbox{Recursive Language}$$

Having seen that a full problem solver needs
an infinite tree-structured syntax and a functional semantics,
we will argue that
we evolved syntax before evolving functional semantics.
If this is true, then
syntax was not the end of the evolutionary road,
 but the beginning of the second leg.
From that point, the road went through
all the components that are needed to build
a functional semantics, until it finally reached recursion.
Investigating the impact of recursion on evolution,
we will find that creativity explodes
when Turing completeness is achieved
causing an evolutionary singularity.
So we agree with Chomsky that syntax was instrumental
in creating the hiatus
that separates our own species from the rest,
but for us the cause of the hiatus is Turing completeness.

And now we can answer the remaining part of our question:
Why did {\it only\/} we humans evolve Turing completeness?
Because Turing completeness is far away,
and only our ancestors took each and every step
of the long and winding road
from animal communication to recursion.
This answer is not as tautological as it seems to be:
in order to be Turing complete,
a species has to invent the word, and then
it has to invent the sentence,
and yet it is not recursive if it
cannot invent a functional semantics, too.
And only a cooperative species needs a language,
as \cite Tomasello (2008) argues, 
so it seems that ours was the only species that was
complex enough and cooperative enough to evolve
a recursive language.

\section{Syntax}

\subsection{Grammar}

\cite Chomsky (1959) presents a hierarchy of grammars.
A \definition{grammar} of a language is a device that is capable
of enumerating all the language sentences.
And, in this context,
\definition{language} is the (usually infinite) set
of all the valid syntactic sentences.

At the end of {\sc Section~2} in that paper, page 143, we read:
``A {\em type 0 grammar $($language$)$} is one that is
  unrestricted.\P\
  Type 0 grammars are essentially Turing machines''.
At the beginning of {\sc Section~3}, same page, we find two theorems.

\begingroup\everypar{}
\hang{\sc Theorem 1}.\quad
For both grammars and languages,
$\hbox{type 0} \supseteq \hbox{type 1} \supseteq
 \hbox{type 2} \supseteq \hbox{type 3}$.

\hang{\sc Theorem 2}.\quad
Every recursively enumerable set of strings is a type 0
language\\(and conversely).

Then {\sc Theorem~2} is explained:
``That is, a grammar of type 0 is a device with the
generative power of a Turing machine.''
\endgroup

From the two theorems we can deduce three corollaries.

\begingroup\everypar{}
\hang{\sc Corollary 1}.\quad
The set of all type 0 grammars (languages) is equal to
the set of all grammars (languages).\newline
This is because, according to {\sc Theorem~1},
type 0 is the superset of all grammars (languages),
and more generally because type 0 is unrestricted.

\hang{\sc Corollary 2}.\quad
For each Turing machine there is a type 0 grammar (and conversely).\newline
This is equivalent to {\sc Theorem~2},
but in terms of grammars (devices) instead of languages (sets).

\hang{\sc Corollary 3}.\quad
For each Turing machine there is a grammar (and conversely).\newline
This results by applying {\sc Corollary~1} to {\sc Corollary~2}.
\par\endgroup

\subsection{Chomsky's Identity}

The third corollary shows that
`Turing machine' and `grammar' are equivalent devices, by which
`computing' and `syntax' are equivalent concepts.
Computing is equivalent to syntax, rather than to language,
because language in \cite Chomsky (1959) refers only to syntax;
it does not refer to semantics, because meanings are not considered,
nor to pragmatics, because intentions are not considered.
Summarizing it, we get Chomsky's identity between
syntax and computing.
$$\hbox{Syntax} = \hbox{Computing}$$

So now we will provide the definition of syntax
that follows from \cite Chomsky (1959)
after adjusting for this deviation towards language:
\definition{syntax} consists of
transformations of strings of symbols,
irrespective of the symbols meanings,
but according to a finite set of well defined rules;
so well defined as a Turing Machine is.
This definition of syntax is very general and
includes natural languages syntax,
just replacing symbols with words and strings with sentences,
but it also includes natural languages morphology or phonology,
taking then sub-units of words or sounds.

As an aside, please note that
this definition of syntax manifests,
perhaps better than other definitions,
what we will call the \definition{syntax purpose paradox}:
syntax, being literally meaningless, should be purposeless.
It is easy to underestimate the value of some
mechanical transformations of strings of symbols.

\subsection{Decidability}

Concerning decidable languages,
\cite Chomsky (1959) states
the following theorem, which is also in page 143.

\begingroup\everypar{}
\hang{\sc Theorem 3}.\quad
Each type 1 language is a decidable set of strings.
\newline
But not conversely, as note 7a adds.
Therefore, $\{{\cal T}_1\} \subset \{{\cal L}_D\}$.
\par\endgroup

The hierarchy of \cite Chomsky (1959) can be extended
to include mildly context sensitive languages in it.
We follow \cite Stabler (2014),
see {\bf Theorem 1} in page 167,
to locate mildly context sensitive languages
in the hierarchy.
$$
\halign{#\hfil&&\enspace#\hfil\cr
\multispan2\bf Chomsky\hfil&\bf Language&\bf Device\cr
\noalign{\smallskip}
Type 0& ${\cal T}_0$& Unrestricted& Turing machine\cr
---& ${\cal L}_D$& Decidable& ---\cr
Type 1& ${\cal T}_1$& Context sensitive& Linear bounded automaton\cr
---& ${\cal L}_M$& Mildly context sensitive& ---\cr
Type 2& ${\cal T}_2$& Context free& Push-down automaton\cr
Type 3& ${\cal T}_3$& Regular& Finite state automaton\cr}
$$
The extended hierarchy is then:
$\{{\cal T}_3\} \subset \{{\cal T}_2\} \subset
 \{{\cal L}_M\} \subset \{{\cal T}_1\} \subset
 \{{\cal L}_D\} \subset \{{\cal T}_0\}$.

The languages defined by minimalist grammars are
mildly context sensitive, and then it seems that
natural language is in this class,
 as argued by \cite Stabler (2014).
This would guarantee that
natural language generation and parsing
are free of infinite loops, as they should be,
because mildly context sensitive languages are decidable,
$\{{\cal L}_M\} \subset \{{\cal L}_D\}$.

\section{Problem Solving}

\subsection{Turing's Identity}

The connection between computing and problem solving
goes back to the very foundation of computing.
\cite Turing (1936) defines his machine in order to prove
that the {\it Entscheidungsproblem},
 which is the German word for `decision problem',
is unsolvable. After defining the Turing machine,
he shows that there is not any Turing machine
that can solve the problem.
This proof is valid only under the assumption
that the set of Turing machines exhausts
the ways of solving problems,
where each Turing machine is a way of solving,
because then that no Turing machine solves a problem
implies that there is no way of solving it.
Summarizing it, we get Turing's identity between
problem solving and computing.
$$\hbox{Problem Solving} = \hbox{Computing}$$

It is even more explicit \cite Post (1936),
who writes that he has in mind a {\it general problem},
and then he goes on to define a `problem solver' which
is very similar and equivalent to a Turing machine.

From then, the identity between computing and
problem solving holds, even if it is not expressed.
For example, for \cite Simon \& Newell (1971)
a theory of human problem solving gets credit
only if a computer program simulates precisely
the behavior of a person solving the same problems.

In Subsection \ref{Chomsky's Identity},
we saw that syntax is computing, and now
we see that problem solving is computing, too.
That both syntax and problem solving are computing
is a coincidence that asks for an explanation
that we will postpone until \ref{Syntax and Problems}.
$$\hbox{Syntax}=\hbox{Computing}=\hbox{Problem Solving}$$

\subsection{Turing Completeness}

After assimilating that problem solving is computing,
the next step is to assume that solving more problems
provides more survival possibilities, and therefore
that the evolution of computing was driven by
the evolution of problem solving, see \cite Casares (T).
In addition, if problem solving is computing, then
the maximum problem solving power is
the maximum computing power, and therefore
the last step in this evolutionary thread
is to achieve the maximum computing power,
which is to achieve Turing completeness.

By definition,
a device is \definition{Turing complete}
if it can compute anything
that any Turing machine can compute.
More informally, a device is Turing complete if
it can be programmed to do any computation,
and then `Turing completeness' is a technical phrase for
`full programmability'.

The Turing machine,
as it was presented by \cite Turing (1936) himself,
models the calculations done by a person
with a finite internal memory who can access
as much external memory as he needs,
meaning that 
we persons can compute whatever a Turing machine can compute
provided that
we can access as much external memory as we need and that
we have enough time to accomplish the computation.
These two conditions refer to memory access and to available time,
and they do not refer to computing capacity,
and therefore
we are Turing complete in computing capacity.
In other words, we are Turing complete because
we can compute any program, that is, because
we can follow any finite set of well-defined rules.

That we are Turing complete is the fundamental
point made by \cite Turing (1936).
For example, according to \cite Davis (1982),
only Turing's model convinced G\"odel
that computing exhausts what is effectively calculable.

Turing completeness is the maximum syntactic power.
This follows directly from
the identity between syntax and computing
seen in Subsection~\ref{Chomsky's Identity}.
We will call a syntax generated by a Turing complete grammar
a \definition{complete syntax}.

The prototype of Turing complete device is
a universal Turing machine, also defined by \cite Turing (1936),
but there are others,
each one associated with one version of recursion.

\section{Recursion}

\subsection{Versions of Recursion}

There are three classical versions of recursion:
\begin{itemize}
\item Recursive functions,
 due to G\"odel, Herbrand, and Kleene,
  where a recursive function
  can be defined without reference restrictions,
  and, in particular, the definition of a function
  can refer to itself.
 Its classical reference is \cite Kleene (1936a).
\item Lambda functions,
 due to Church, Kleene, Rosser, and Curry,
 which are higher-order functions, that is,
  functions on functions to functions.
 In lambda-calculus everything is a function,
 or a variable.
 Its classical reference is \cite Church (1935).
\item Computable functions,
 due to Turing, and Post,
 that are ruled transformations of strings of symbols.
 There is a universal computer
 for which any computer is a string of symbols.
 Its classical reference is \cite Turing (1936).
\end{itemize}
There are now more versions of recursion,
but here we will manage with these three.

Although apparently very different to each other,
the three are equivalent:
\begin{itemize}
\item \cite Kleene (1936b)
showed that every lambda function is recursive, and the converse.
This means that for each lambda function
there is a recursive function that performs the same calculation,
and the converse, and therefore
lambda functions are equivalent to recursive functions.
\item \cite Turing (1937) showed that
every lambda function is computable,
and that every computable function is recursive.
This means that
for each lambda function there is a Turing machine
that performs the same calculation,
and that
for each Turing machine there is a recursive function
that performs the same calculation,
and therefore, together with \cite Kleene (1936b),
computable, lambda, and recursive functions
are equivalent.
$$\hbox{Computable Function} = 
  \hbox{Lambda Function} = 
  \hbox{Recursive Function}$$
\end{itemize}

These mathematical equivalences show that
we can implement recursion in several ways.
Each implementation has its own
advantages and disadvantages,
and then they are not completely equivalent
from an engineering point of view.
So let us now introduce the devices
that implement these kinds of functions:
\begin{itemize}
\item The universal Turing machine,
 already seen in \ref{Turing Completeness},
is a Turing complete device because it can
compute whatever a Turing machine can compute,
and then it can calculate any computable function.
And also, by the equivalences between functions,
a universal Turing machine can calculate
any lambda and any recursive function. 
\item By definition, a lambda-calculus interpreter can
calculate any lambda function, and then,
by the equivalences,
it can calculate any computable and any recursive function.
\item Finally, an ideal (error free and eternal) mathematician
can calculate any recursive function,
and then any lambda and any computable function.
This implementation of genuine recursion
can be called formal mathematics.
\end{itemize}

The precise conclusion is that
whatever a universal Turing machine can calculate,
it can also be calculated by a lambda-calculus interpreter,
and by a mathematician using recursive functions,
and the converse of both.
Therefore, we say that
a universal Turing machine,
a lambda-calculus interpreter,
a mathematician calculating recursive functions,
or any other mathematically equivalent calculating device,
is a \definition{universal computer}.
Only universal computers implement recursion. Then
`universal computer' is synonymous with
`Turing complete device', and also with
`recursive device'.
$$\hbox{Universal Computer} = 
  \hbox{Turing Complete Device} = 
  \hbox{Recursive Device}$$

\subsection{Properties of Recursion}

Recursive devices have some properties in common:
incompleteness, extendability, and undecidability,
all three summarized by \cite Post (1944).
The three properties are related, but here we are
most interested in one: undecidability.
A calculating device is \definition{decidable}
if it can resolve, usually with a yes-or-no answer,
in every case, that is, for any finite input data.
Otherwise, it is undecidable.
Then an undecidable device can get stuck
in an infinite loop, never ending a calculation.
As we can program a Turing complete device
to do any computation, we can program it 
to do an infinite loop, and therefore
every recursive device is undecidable.

Note that these properties are properties of devices,
and not properties of functions.
A recursive function can be decidable,
but all recursive devices are undecidable.
This can be a source of confusion,
because only the most capable devices are recursive,
while even the simplest functions are recursive.
An example of simple recursive function is
the identity function that just returns what it takes,
and then it is a `no operation' instruction
that, in fact, does not compute anything.
To prevent this confusion,
except when we refer explicitly to recursive functions,
we will always use recursion, or recursive,
to refer to the property of devices.

Warning note:
We are using here the most exacting
definition of recursion possible,
equating recursion to Turing completeness.
Though this severe definition of recursion prevents confusion
and succeeds where other definitions fail,
mainly in discriminating human versus non-human capacities,
see \cite Casares (R),
less demanding definitions of recursion pervade linguistics,
see \cite Stabler (2014)
and \cite Watumull et al. (2014), for example.
Different purposes demand different definitions,
but this should not be a problem provided
we are always aware of which definitions are ruling.
And here, recursion is Turing completeness.

The existence of different implementations of recursion
can be another source of confusion.
Although from a mathematical point of view
every computable function is recursive, and the converse,
it is sometimes said that a function is recursive
when it was implemented using self-reference, but not
when the same function was implemented using iteration,
for example.
Here, we will avoid that loose way of speaking, because,
for example, the identity function, which does not refer to itself,
is nevertheless genuinely recursive,
see \cite Kleene (1936a), page 729, where it is called $U_1^1$.

So recursive devices are undecidable,
and then if you can prove that a device,
whatever the arguments it takes,
always completes the calculations in a finite time,
then you have proved
that the device does not implement recursion,
or, in other words,
that it is not Turing complete.
The converse is not true;
that a computation does not complete 
does not imply that the computer is recursive.

\subsection{Recursive Language}

A recursive device can calculate any recursive function,
but, to do it, it needs the expression of the function.
Then, each universal computer understands a way
to express any possible recursive function.
We will call
this language that every Turing complete device
needs to express any recursive function
a \definition{recursive language}.
In other words,
the recursive language is the language used
to program the universal computer.
Note that, in \cite Casares (H) and everywhere else,
this kind of language is called a \definition{Turing complete language},
or a \definition{complete language} ${\cal L}_C$, for short.

For example, a typical lambda-calculus interpreter
understands a recursive language~$\Lambda$ with a syntax
defined inductively this way:
$$\eqalign{
x \in V &\quad\;\; V \subset \Lambda\cr
x \in V &\Rightarrow x' \in V\cr
x \in V,\; M \in \Lambda &\Rightarrow (\lambda x\, M) \in \Lambda\cr
M,N \in \Lambda &\Rightarrow (M\, N) \in \Lambda\cr
}$$
This means that 
in lambda-calculus there are only:
an infinity of variables built by priming,
 which belong to the subset $V$;
lambda abstractions,
 that is, function definitions, $(\lambda x\, M)$; and
function applications, $(M\, N)$.
These last two are ordered pairs
that grow into binary trees by induction.

Other universal computers use other syntaxes,
so implementing binary trees is not a requirement for recursion.
For example, a typical universal Turing machine understands the
transition table of the Turing machine to emulate.
That table is a list of conditionals:
if the current internal state is $q_i$,
 and the symbol read is $s_r$ ($s_r$ can also be {\it blank\/}), then
 go to internal state $q_n$ ($q_n$ can be equal to $q_i$),
 write symbol $s_w$ ($s_w$ can be equal to $s_r$ or not,
  even {\it blank\/}),
 and move to the {\it right}, or to the {\it left}, or {\it halt}.
In fact, as explained in \cite Casares (U),
the only syntactic requirement
that any recursive language has to fulfill is
that the set of its syntactic objects
has to be infinite enumerable or bigger, that is,
any infinite set complies.

While these Turing complete devices implement 
complete syntaxes, as defined in \ref{Turing Completeness},
they need also semantics.
Syntax alone is not enough, and some semantics are needed.
It is not enough that the expressions represent the functions,
the Turing complete device has to understand what
the expressions mean to produce
the results that the represented functions produce.
Here, we will call any semantics providing the meaning
of recursive functions a \definition{functional semantics}.
And then, a functional semantics is needed
to implement a complete syntax!

For example, the typical lambda-calculus interpreter
has to know how to perform $\beta$-reductions,
see \cite Curry \& Feys (1958), \S3D,
as for example $((\lambda x\, x)\, x') \rightarrow x'$,
in order to be able to calculate any lambda function.
Then, $\beta$-reduction is part of the functional
semantics that makes this interpreter a recursive device.
Note that $(\lambda x\, x)$ is the identity function,
$U_1^1 = (\lambda x\, x)$.

Complete syntax is synonymous with recursion, and
a somewhat paradoxical property of recursion
is apparent when we call it complete syntax.
Complete syntax is syntax,
 as defined in \ref{Chomsky's Identity},
but it is also more than syntax,
 because complete syntax requires a functional semantics.
To solve the paradox,
we will consider functional semantics part of semantics
and also part of syntax, 
 because functional semantics only provides the
 meanings of functions, and functions are the core of syntax.
This way, functional semantics is in the intersection
of semantics and syntax.

\subsection{Universal Grammar Is Universal}

Decidability partitions the set of grammars,
and the set of the corresponding languages, in two subsets.
For languages we get:
$\{{\cal L}_D\} \cap \{{\cal L}_{\!\bar D}\} = \emptyset$ and
$\{{\cal L}_D\} \cup \{{\cal L}_{\!\bar D}\} = \{{\cal T}_0\}$.
The hierarchy of \cite Chomsky (1959) is in
the decidable part of $\{{\cal T}_0\}$,
see \ref{Decidability}, while this other hierarchy
is in the undecidable part of $\{{\cal T}_0\}$.
In \cite Casares (H), both parts are combined
and extended into a complete hierarchy.
$$
\halign{#\hfil&&\enspace#\hfil\crcr
\multispan2\bf Chomsky\hfil&\bf Language&\bf Device\cr
\noalign{\smallskip}
Type 0& ${\cal T}_0$& Unrestricted& Turing machine\cr
---& ${\cal L}_{\!\bar D}$& Undecidable& ---\cr
---& ${\cal L}_C$& Recursive& Universal Turing machine\cr}
$$
The undecidable hierarchy is then:
$\{{\cal L}_C\} \subset \{{\cal L}_{\!\bar D}\} \subset \{{\cal T}_0\}$.

But this hierarchy is misleading
because of the peculiar nature of universal Turing machines.
A single universal Turing machine
is as powerful as the set of all Turing machines,
but it needs a recursive language to program it.
Then, we should distinguish three languages
when dealing with universal computers:
the recursive language $L_C[L_S]$
that uses a syntax $L_S$
to describe the programmed language $L_P$.
We can write this as $L_C[L_S](L_P)$,
meaning that $L_C[L_S] \in  \{{\cal L}_C\}$
is implemented in the universal computer hardware,
and $(L_P)$, where $L_P \in \{{\cal T}_0\}$, is programmed software.
Then, in our case,
the universal grammar of \cite Chomsky (1965) implements $L_C[L_S]$,
and $L_P$ is any language that we can learn,
say English or {\sc Lisp}.

The equation of the recursive languages (1)
is a consequence of the Turing completeness of its grammar,
and it states that we can produce any language with
a recursive language. In mathematical terms,
$$\forall L_P \in \{{\cal T}_0\},\;
\forall L_C[L_S] \in \{{\cal L}_C\},\quad
L_C[L_S](L_P) = L_P
\;. \leqno(1)$$

The equation of the recursive languages (1) explains
why we can learn any natural or artificial language.
And it also explains why the universal grammar does not
set any specific language.
So it is not because the universal grammar is underspecified,
as proposed by \cite Bickerton (2014),
but because the universal grammar is universal,
in the sense that it is a universal computer.
Then, the universal grammar is overqualified,
rather than underspecified.
This subject is further elaborated in
\cite Casares (U).

Let us now differentiate the two components $L_C$ and $L_S$
of the recursive language $L_C[L_S] \in \{{\cal L}_C\}$.
First, we will see that $L_S$ can be a decidable language,
that is, that $L_S \in \{{\cal L}_D\}$ is possible.
For example, the syntax of the recursive language of
the typical lambda-calculus interpreter is a context free language
that uses six terminals, 
$V_T = \{ x, ', (, \lambda, \sep, ) \}$,
where \sep\ is a space,
two non-terminals, $V_N = \{ S, X \}$,
and that obeys the following five generative rules.
$$\eqalign{
S &\rightarrow X \mid (\lambda X \sep\, S) \mid (S \sep\, S)\cr
X &\rightarrow x \mid X'}$$
Context free languages generate trees,
but the syntax of a recursive language can be simpler.
For example, the syntax of the recursive language of
the typical universal Turing machine is a regular language,
because any Turing machine can be expressed as
a list of five-word clauses $QY\!QY\!Z$,
where $Q=\{q_0,q_1,\dots,q_{_Q}\}$ is the finite set of internal states,
$Y=\{ \hbox{\it blank}, s_1, \dots, s_{_Y}\}$
 is the finite set of symbols, and
$Z=\{ \hbox{\it left},\hbox{\it right},\hbox{\it halt}\,\}$
 is the set of movements, and then
its syntax is the Kleene closure of
 the five-word clauses, $(QY\!QY\!Z)*$.
Second, we know that any recursive language $L_C[L_S]$
is necessarily undecidable
because it can be programmed to generate any language,
including undecidable languages.
This shows that
$$\exists L_S \in \{{\cal L}_D\},\quad
 L_C[L_S] \in \{{\cal L}_C\} \subset \{{\cal L}_{\!\bar D}\}
\;. \leqno(2)$$

In those cases where $L_S \in \{{\cal L}_D\}$,
only functional semantics can provide the resources
needed to achieve universality,
so we will distinguish the proper syntax defined by $L_S$
from the functional semantics that we ascribe to $L_C$.
In other words,
the recursive language $L_C[L_S]$ of the recursive device
has a syntax $L_S$ and a semantics $L_C$.
Equation (1) when $L_S \in \{{\cal L}_D\}$ shows that
it is possible to program any language $L_P$
using a decidable syntax $L_S$ and
a suitable functional semantics $L_C$.
In mathematics we say that
from (1) and (2) we deduce (3):
$$\forall L_P \in \{{\cal T}_0\},\;
  \exists L_S \in \{{\cal L}_D\},\quad
  L_C[L_S](L_P) = L_P
\;. \leqno(3)$$
The conclusion is that
the syntax of a recursive language can be decidable.
Then it should be decidable.
And in our case, it is decidable,
or so it seems as seen in \ref{Decidability}.

\subsection{The Key}

The equation of the recursive languages (1)
shows that to implement one specific language
we can either build a recursive device and
then program the specific language
in its recursive language,
or instead build the specific grammar for the language.
To note that the second is much easier than the first
is to rephrase the anomaly of syntax.

On the other hand,
once a recursive device is available,
it would not be wise to limit it.
Then every particular natural language
should be a recursive language
with its functional semantics.
We are saying that,
when the $L_P$ of the equation (1)
is a particular natural language,
then $L_P \in \{{\cal L}_C\}$
should be the case.
We know it is possible,
because $L_P$ can be any language,
and failing $L_P$ to be recursive
would imply that we could not
express and mean every possible
recursive function in that
particular natural language $L_P$.

Is there any particular natural language that is not recursive?
\cite Everett (2008) claims that the Pirah\~a language
lacks recursion.
I don't know, but if it is true,
then you cannot translate a manual of {\sc Lisp}
from English to Pirah\~a,
as for example \cite McCarthy et al. (1962).

Another simpler test concerns logical paradoxes.
Logical paradoxes lay completely on functional semantics,
not needing any other semantics.
For example, take the English sentence~(4).
\begin{flushleft}
(4)\quad This sentence is false.
\end{flushleft}
The syntactic analysis of (4) is straightforward,
but when we look for its meaning
using English functional semantics
we get trapped in an infinite loop.
Now, the test:
If it is not possible to translate this sentence (4) to Pirah\~a,
then Pirah\~a is not recursive.
This test is easier than that of translating a {\sc Lisp} manual,
but it is not as conclusive, because
if the translation of the sentence (4) were possible,
then nothing would be settled.

Three notes before leaving the Pirah\~a language.
\begin{itemize}
\item Although unexpected,
a non-recursive particular natural language is possible,
because the $L_P$ of the equation of the recursive languages (1)
can be any language.
\item The syntax of a recursive language
does not need to use tree data structures,
as shown by the syntax of 
the typical universal Turing machine recursive language,
which only uses lists of fixed length clauses,
so the Pirah\~a language can still be recursive.
The question is whether the functional semantics
that complement the simple Pirah\~a syntax is enough or not.
\item Even if the Pirah\~a language were not recursive,
the Pirah\~a speakers could still learn English,
and then they would understand sentence (4)
and a {\sc Lisp} manual in English.
\end{itemize}

Going back to our main business,
we were seeing that,
while any recursive language is undecidable,
its syntax can be, and should be, decidable.
This explains an otherwise paradoxical fact:
our syntactic abilities are in excess to those needed
to speak a natural language.
We are Turing complete,
 which is the maximum syntactic capacity,
 as required for better problem solving,
though the natural language syntax parsers are not Turing complete,
 to assure parsing completion.
In other words,
recursion is not a requirement of natural language syntax,
and being recursion a syntactic property, we conclude that
recursion is not a requirement of language.

We will repeat it just because this is the key of the paper;
once this is assimilated,
everything else is pretty straightforward.
Both syntax and problem solving are computing, but
while recursion,
 which is the property that signals the maximum computing power,
is detrimental to any syntax,
 even to the syntax of a recursive language,
the very same recursion is the most desirable property
for problem solving.

\subsection{Human Uniqueness}

Because parsing tree data structures
does not require Turing completeness, then
 a species that is not Turing complete
 can use tree data structures.
On the other hand,
it seems that we are the only Turing complete species,
that is, the only one implementing recursion.

To prove that a device is Turing complete
you have to show that it can compute any recursive function.
Therefore, to be qualified as Turing complete,
a device has to take
 the expression of any recursive function in some language,
 and any data,
and then return the right result every time.
This implies that
 a recursive species needs a recursive language.
The only known species with a recursive language is ours, 
and then it seems that we are the only recursive species.

In any case,
the equation of recursive languages
applies to any Turing complete device,
and this implies that the members of any recursive species,
whether they use a recursive language to communicate or not,
can learn any language,
as it happens to the Pirah\~a speakers
independent of the recursivity of their language.
We are saying that the members of any recursive
species can learn English, {\sc Lisp},
and in fact any language,
and that that capability is independent of the interfaces,
a conclusion that explains why
our spoken language can be written
without any further evolutionary intervention.

If we are in fact the only Turing complete species,
then recursion is uniquely human.
It is also true that
 any human organ is also uniquely human,
 that is, the human eye is not like any other,
and an anthropologist can distinguish a human bone
from any non-human bone.
But perhaps recursion is different,
because it is a property that 
our species {\it Homo sapiens\/} has,
and every other species has not,
and because recursion is a singular property,
as explained below in Subsection~\ref{Problems and Evolution}.

If recursion is a requirement of problem solving,
 and by this we mean that 
 recursion has some survival value for problem solving,
 which is very likely true
 because of the recursive nature of problems
 (see Section~\ref{Problem Solving}),
then
 the hypothesis that `what is uniquely human and
 unique to the faculty of language ---the faculty of language
 in the narrow sense (FLN)--- is recursion',
by \cite Hauser, Chomsky, and Fitch (2002),
is only partly true.
While our species seems to be the only Turing complete one,
the recursive machinery is being used not only by language
but also by problem solving.

Here, instead of the complete and ambiguous formulation
of the hypothesis in page 1573,
``FLN comprises only the core computational mechanisms of recursion
as they appear in narrow syntax and the mappings to the interfaces'',
we are using the simpler formulation `FLN is recursion'
for three reasons:
\begin{itemize}
\item `FLN is recursion' is shorter and it was used twice
by \cite Hauser, Chomsky, and Fitch (2002),
in the abstract and in the very same paragraph where the longer one is.
\item Language can be spoken, written, and signed, and then it should
be independent of the interfaces, or else it had to evolve three times.
\item The mappings to the interfaces are irrelevant here, because
our point is that recursion is not used only by language,
and this negates both formulations of the hypothesis.
\end{itemize}

\subsection{Our Hypothesis}

Even if the `FLN is recursion' hypothesis
is not completely true,
it touches one of the most important questions
about language and syntax: their relation to recursion.
To fix that hypothesis, we dare to state a new one
that takes into account problem solving.
Our hypothesis is that
`syntax and problem solving co-evolved in humans
towards Turing completeness'.

The argument goes like this:
Solving more problems provides more survival opportunities,
and then, because of the recursive nature of problems,
this thread of evolution goes naturally towards Turing
completeness (Section~\ref{Problem Solving}).
And Turing completeness requires a recursive language
in which to express and mean any recursive function
(Subsection~\ref{Recursive Language}).
Then our syntax should have evolved to satisfy the requirements
of problem solving.

To be clearer,
we agree with Chomsky and his `FLN is recursion' hypothesis
in the following points:
\begin{itemize}
\item we are the unique species with a recursive language,
\item we are the unique recursive species,
\item recursion is a syntactic property, and
\item recursion is not a requirement of syntax,
          so it is not a requirement of language.
\end{itemize}
But then Chomsky does not want to go further.
In fact, he cannot go further because,
if recursion is not a requirement of language,
 meaning that recursion has not survival value for language,
and recursion is exclusively a language property,
 as stated by his `FLN is recursion' hypothesis,
then recursion cannot have an evolutionary explanation.
In the next subsections,
 \ref{Merge Is Not Recursion} and
 \ref{Chomsky on Syntax Evolution},
we will try to understand why Chomsky reached this conclusion.

Additionally, for us,
\begin{itemize}
\item recursion is a requirement of full problem solving,
\end{itemize}
and then recursion can have an evolutionary explanation,
and we can state our hypothesis:
`syntax and problem solving co-evolved in humans
 towards recursion'.

\subsection{Merge Is Not Recursion}

\cite Chomsky (2007), page 20,
is wrong when he writes:
``If the lexicon is reduced to a single element,
then unbounded Merge will easily yield arithmetic.''
The offending word is `easily'.
The definition of Merge, in the same page, is:
$$\hbox{Merge}(X,Y) = \{ X, Y \} \;.$$
\cite Chomsky (2006), page 184, states it more
precisely:
``The most restrictive case of Merge
applies to a single object, forming a singleton set.
Restriction to this case yields the successor function,
from which the rest of the theory of natural numbers
can be developed in familiar ways.''
The theory of natural numbers is also known as arithmetic.

Let us call the single element of the lexicon $\emptyset$.
Then,
$$\hbox{Merge}(\emptyset) =
  \hbox{Merge}(\emptyset,\emptyset) =
  \{ \emptyset, \emptyset \} =
  \{ \emptyset \} \;,$$
which is indeed a singleton set.
Reiterating we obtain:
$\hbox{Merge}(\hbox{Merge}(\emptyset)) =
 \hbox{Merge}(\{\emptyset\}) =
 \{\{\emptyset\}\}$,
$\hbox{Merge}(\hbox{Merge}(\hbox{Merge}(\emptyset))) =
 \hbox{Merge}(\{\{\emptyset\}\}) =
 \{\{\{\emptyset\}\}\}$, and so on.
In the construction of the natural numbers
by \cite Zermelo (1908):
$0 = \emptyset$,
$1 = \{\emptyset\}$,
$2 = \{\{\emptyset\}\}$,
$3 = \{\{\{\emptyset\}\}\}$,
and so on. Now, rewriting,
$\hbox{Merge}(0) = 1$, 
$\hbox{Merge}(1) = 2$, 
$\hbox{Merge}(2) = 3$,
$\hbox{Merge}(n) = n+1$,  
and then Merge is indeed the successor function.
So far, so good.

But the successor function is not enough to
develop the theory of natural numbers.
This question was answered by \cite Kleene (1952)
in his exhaustive investigation
of the genuine recursion of G\"odel.
The successor function is only Schema I,
the simplest of the six (I-VI) schemata that are needed
to define any recursive function of natural numbers,
see pages 219 and 279.
For example, Schema III are the identity functions,
$U_i^n$, which are also needed to implement arithmetic.

The lambda version of recursion
uses a version of Merge that we will call {\tt cons}.
As Merge, {\tt cons} takes two arguments and returns
an object that contains the arguments, and the only
difference is that the result of {\tt cons} is
an ordered pair, instead of a set:
$$\hbox{\tt cons}(X,Y) = (X\sep\, Y) \;.$$
By iterating uses of {\tt cons},
we can build binary trees with left and right branches,
while using Merge we cannot distinguish branches.
Again, {\tt cons} alone is not enough
to implement arithmetic in the lambda version of recursion,
because $\beta$-reductions and variables are also needed.

This is as it should be.
Merge is decidable,
that is, a device implementing only Merge
will always complete its calculations in a finite time,
and therefore a device implementing only Merge
is not recursive.
By the way, the same happens to
{\tt cons} and to the successor function.
It is as it should be because,
being decidable, Merge meets the requirement
of syntax parsers,
and then Merge is well suited to be
the main operation of syntax,
as proposed by the minimalist program.

On the other hand, arithmetic is undecidable.
This is arguably the most fundamental truth of mathematics:
in any finitary formalization of
arithmetic that is consistent
there are arithmetic statements
that cannot be decided true or false
within the formalization,
proved by \cite G\"odel (1930) for genuinely recursive formalisms,
by \cite Church (1935) for lambda formalisms,
and by \cite Turing (1936) for computable formalisms,
where a finitary formalization is an implementation.
Recursion was the result of this
paramount investigation on the foundations
of mathematics.

The chasm that separates the single successor function
from a full implementation of recursion
shows us the anomaly of syntax from a new perspective.
The most we could say would be that
Merge is required to implement recursion,
but even this is not completely exact.
A version of Merge, {\tt cons}, is needed to
implement the lambda version of recursion,
and another version of Merge, the successor function,
is needed to implement the genuine version recursion,
but the computable version of recursion
only requires the conditional,
 and some auxiliary operations not related to Merge,
to implement its functional semantics.
In summary: Merge is not recursion,
it is only an element that can be used
to implement recursion. And, in any case,
the way from Merge to recursion is not easy;
it takes some steps, and not all are easy.

\subsection{Chomsky on Syntax Evolution}

Now, we will try to understand Chomsky
on the evolution of syntax and recursion,
under the light of the anomaly of syntax.

Presenting the `FLN is recursion' hypothesis,
\cite Hauser, Chomsky, and Fitch (2002)
wrote: ``If FLN is indeed this restricted,
this hypothesis has the interesting effect of
nullifying the argument from design, and thus
rendering the status of FLN as an adaptation
open to question'', page 1573.

\cite Pinker \& Jackendoff (2004) disagreed:
``Our disagreement specifically centers on the hypothesis
that recursion is the only aspect of language that is special to it,
that it evolved for functions other than language, and
that this nullifies `the argument from design' that sees
 language as an adaptation'', page 205.
Then, in pages 216--217, arguing ``that unlike humans, 
tamarins cannot learn the simple recursive language $A^n B^n$'',
Pinker \& Jackendoff state the anomaly of syntax:
``If the conclusion is that
human syntactic competence consists only
of an ability to learn recursive languages
(which embrace all kinds of formal systems,
including computer programming languages,
mathematical notation, the set of all palindromes,
and an infinity of others),
the fact that actual human languages are a minuscule
and well-defined subset of
recursive languages is unexplained.''

It was in fact unexplained by Pinker \& Jackendoff,
and also by
\cite Fitch, Hauser, and Chomsky (2005) in their answer, who
on this point wrote:
``Fitch \& Hauser do not even mention recursion in the cited
paper, and the generation of limited-depth hierarchical phrase
structure was not confused with recursion in that paper
(although it was by some commentators on the article)'', page 204.
Of course, they are twice right:
recursion is frequently confused, and
even unbounded $A^n B^n$ is not a recursive language
because it is not possible to calculate
every recursive function in a language
that has not a functional semantics.
The salient point, however, is that
they wrote nothing about the anomaly.

But \cite Chomsky (2006) himself was indeed aware
 of the anomaly of syntax.
We copy here the complete paragraph between pages
184 and 185:
\begin{quotation}
The most restrictive case of Merge applies to a single object,
forming a singleton set. Restriction to this case yields the
successor function, from which the rest of the theory of natural
numbers can be developed in familiar ways. That suggests a
possible answer to a problem that troubled Wallace in the late
nineteenth century: in his words, that the ``gigantic development
of the mathematical capacity is wholly unexplained by the theory
of natural selection, and must be due to some altogether distinct
cause,'' if only because it remained unused. One possibility is
that the natural numbers result from a simple constraint on the
language faculty, hence not given by God, in accord with Kronecker's
famous aphorism, though the rest is created by man, as he continued.
Speculations about the origin of the mathematical capacity as
an abstraction from linguistic operations are not unfamiliar.
There are apparent problems, including dissociation with lesions
and diversity of localization, but the significance of such
phenomena is unclear for many reasons (including the issue of
possession vs.\ use of the capacity). There may be something to
these speculations, perhaps along the lines just indicated.
\end{quotation}
Perhaps not, but it seems that for Chomsky
our mathematical capacity, which includes recursion,
``is wholly unexplained by the theory of natural selection,
[\dots] if only because it remained unused.''

Our conclusion is that Chomsky took the anomaly of syntax
as a proof that recursion was not selected,
and then he proceeded consequently.
For \cite Chomsky (2005a),
not too long before about $50\,000$ years ago,
a mutation rewiring the brain took place
yielding the operation of Merge.
Because of some causes not related to natural selection,
also called third factor causes by \cite Chomsky (2005b),
Merge had the effect of producing {\it easily\/}
the complete recursion;
but remember \ref{Merge Is Not Recursion}.
Then,  according to \cite Chomsky (2000), page 163,
 where he argues that selection is
 ``[a] factor, not {\it the\/} factor'',
recursion could be
the emergent product of the spontaneous self-organization
of complex systems under the strictures of physical law.
In summary: Chomsky had to invoke third factor causes
to explain recursion,
instead of invoking evolutionary or first factor causes,
because he took the anomaly of syntax at face value
and he is consequent. 

Our position differs from that of Chomsky:
we can explain the survival value of recursion, and then
recursion is not an incidental result of language.
Under our hypothesis, recursion was selected
because of its problem solving advantages,
despite its syntactic inconvenience.
Then, Wallace problem is not a problem, and
we were using recursion for solving problems
since we achieve Turing completeness,
though its mathematical formalization came only recently.
And, for us, see \ref{Merge Is Not Recursion},
the way from Merge to recursion is not easy,
it takes some steps, and not all are easy.
Then, Merge could be much more ancient,
and it is possibly shared with other species,
because the binary tree data structure
is simple and useful.

\section{Evolution}

\subsection{Requirements}

It is true that we can express any
recursive function in our natural language,
and that by reading the expression
we can understand how the function works,
to the point of producing the right results.
This is just a different way to state that
our natural language is recursive.
On the other hand, problem solving
achieves its maximum power when Turing completeness
is achieved, and Turing completeness
is mathematically equivalent to recursion.
For our hypothesis, this happens because
problem solving and syntax are both computing,
and therefore computing was shaped
by taking evolutionary requirements from both, and then
while Turing completeness is not a requirement of syntax,
our natural language is nevertheless recursive because
Turing completeness is a requirement of full problem solving.
In other words, some traits of language
were selected, totally or partly, because
they enhanced our problem solving abilities
increasing our survival fitness.

To better assess our hypothesis,
in this Section~\ref{Evolution}
we will analyze those requirements.
Our guide to problem solving in this investigation
will be the programming language {\sc Lisp},
as defined originally by \cite McCarthy (1960).
We have chosen {\sc Lisp} for three reasons:
\begin{itemize}
\item {\sc Lisp} has stood the test of time.
{\sc Lisp} is still active in
the field of Artificial Intelligence, where
it makes easy the resolution of many problems.
\item {\sc Lisp} syntax is minimalist,
using only binary trees.
The {\sc Lisp} syntax parser is the simplest
two-way parser.
\item {\sc Lisp} is based on lambda-calculus,
but it fuses the functional semantics
of the three classical versions of recursion.
\end{itemize}
In summary, {\sc Lisp} is a language
well suited for solving problems
that uses the simplest syntax and
implements all versions of recursion.

Although the election of {\sc Lisp} is not arbitrary,
it is neither completely neutral,
and some findings could be an artifact of
choosing {\sc Lisp}.
To counter this possibility as much as possible,
we will argue the need of every feature
that full problem solving requires,
following \cite Casares (P),
and we will explain how each feature
is implemented in {\sc Lisp}.
In summary, we will use {\sc Lisp} to guide
our investigation on the relation
between natural language and problem solving.
You can judge our election
at the end of this Section \ref{Evolution}.

\subsection{Variables}

To express a problem
we have to refer to its freedom and to its condition.
To name the freedom we have to use a word that does not refer
to anything, that is, it has to be a word free of meaning.
For example, if the problem is that we do not know what to do,
then its more direct expression in English is `what to do?'.
In this sentence, the word `what' does not refer to anything
specifically, but it is a word free of meaning and purely
syntactical, that is, a function word.
In particular, if the answer to this question were `hide!',
then the meaning of the word `what' in the question would be `hiding',
but if the answer were `run!', then it would mean `running'.

Frequently we answer a question with a single word and
no additional syntax, which is an indication that
in these cases a bi-sentential syntactic analysis is needed
because the pair question-answer is a grammatical whole,
see \cite Krifka (2011).
For example, the meaning of the pair:\\
\null\qquad (Q) --- Who is reading `War and Peace'?\\
\null\qquad (A) --- Bill.\\
reduces to\\
\null\qquad (M) --- Bill is reading `War and Peace'.\\
where `Bill' is the meaning of `Who'.

In mathematics, the typical way to express the freedom of a problem
is to use the unknown $x$, which works the same way as the
interrogative pronoun `what'. For example, if we want to know
which number is the same when it is doubled as when it is squared,
we would write:
$$x? [2x = x^2] .$$
The condition is the equality $[2x = x^2]$, in this case.
Equality is a valid condition because it can be satisfied, or not.

The $x$ is just a way to refer to something unknown,
so we could use any other expedient
just by indicating it with the question mark (?).
This means that
$$y? [2y = y^2]$$
is exactly the same problem.
We call this equivalence $\alpha$-conversion,
following \cite Curry \& Feys (1958), \S3D.

Natural languages do not provide
inexhaustible collections of unknowns or variables,
as lambda-calculus and {\sc Lisp} and mathematics do,
but only a few wh-terms that are not completely
free of meaning.
As observed by \cite Hamblin (1973),
`who' is not completely free of meaning
because it refers to a person or, in other words,
it refers to an indefinite individual
belonging to the set of persons.
Because of this,
there is not $\alpha$-conversion in natural languages,
and it is very difficult to express and to understand
problems involving several unknowns of the same wh-type
in natural languages.
A consequence is that
a mathematical notation to complement the natural language
is needed.
For us, the difficulty is an indication that,
concerning unknowns or variables,
language was first and problem solving last.

\subsection{Sentence}

It is important to note that
the unknown has to be part of the condition,
in order to determine if a value is a solution to the problem, or not.
In the condition, the unknown $x$ is a free variable,
and therefore the condition is an open expression,
that is, a function.

In summary, in order to refer to the freedom of a problem
we have to use free variables,
which are words free of meaning that do not refer to anything.
These free words are useless by themselves,
we can even substitute one for another using an $\alpha$-conversion,
so they have to be combined with other words
to compose the condition of the problem.
We will call this structure of words a \definition{sentence}.
All things related to the sentence,
as, for example, the rules for word combination,
are what is usually called \definition{syntax}.
So, as separate words are not enough to express problems,
then some structure of words is needed, and therefore
syntax is needed to represent problems.

{\sc Lisp} uses binary trees for its syntax.
In {\sc Lisp}, any expression, technically called S-expression,
is either a word or a sentence,
 where a sentence is an ordered pair.
In {\sc Lisp} the function to generate a sentence
is {\tt cons}, already seen in \ref{Merge Is Not Recursion},
that takes two expressions and returns
an ordered pair. Because the arguments of {\tt cons}
can also be ordered pairs,
the resulting structure is a binary tree.
Then, the result of {\tt (cons X Y)} is a tree where
the expression {\tt X} is the left branch
and the expression {\tt Y} is the right branch.
{\sc Lisp} also provides the corresponding deconstructing
functions: {\tt car}, that takes a tree as argument and returns
the left branch of the tree,
and {\tt cdr}, that returns the right branch.
The analysis of sentences in {\sc Lisp} requires also
a predicate, {\tt atom?}, that returns {\tt t}, for {\sc true},
when it is applied to a word, and {\tt nil}, for {\sc false},
when it is applied to a sentence.
This is the complete apparatus that {\sc Lisp}
needs to generate any sentence from its constituent words,
and to reach the words that are any sentence.
Then, for {\sc Lisp} syntax,
generation is simpler than proper parsing,
but in any case all these operations are decidable,
so the {\sc Lisp} syntax parser is decidable.

The {\sc Lisp} syntax parser is arguably
 the simplest two-way,
word to sentence and sentence to word, parser.
The two simplest composing functions are
{\tt cons} and Merge, because they compose
only two objects, being the only difference that
the result of {\tt cons} is ordered, while
the result of Merge is unordered.
If we use Merge instead of {\tt cons},
then some marker or label, as a preposition or simply stress,
is needed to signal the elements of the sentence.
Markers give more flexibility, because then
it is a pragmatic decision how to order
the information conveyed by a sentence.
On the other hand, when order is prescribed by syntax,
parsing is simpler and needs less memory,
because then the parser always knows what to expect.
In summary: the Merge parser is nearly as efficient
as the optimal {\tt cons} parser,
and much more flexible.

This shows that there is a trade-off between
the parsing efficiency of {\tt cons},
and the pragmatic convenience of Merge.
Some natural languages use order, some markers,
 and most both,
suggesting that evolution has selected
the more flexible Merge as an answer to the trade-off
between syntactic efficiency and pragmatic convenience.
In that case, Merge was selected.

We saw that syntax is needed to express problems, 
and that {\sc Lisp} uses a tree sentence structure,
which is simpler than, but similar to, the natural languages
sentence structure generated by Merge.
What we left pending is the check against
problem solving, that is,
whether problem solving requires
a tree sentence structure, or not.

\subsection{Problem}

We have seen
why we need sentences, that is, syntax,
to express problems,
and why separate words are not enough.
We have also seen two types of word:
semantic or content words with meaning, and
syntactic or function words without meaning.
Well, although it seems impossible,
there is a third type of word: defined words.

A defined word is just an abbreviation,
so we could go on without them, but they are handy.
That way we substitute a word for a whole expression.
It is the same expedient that we use when
we introduce a technical term to encapsulate a phrase
in order to avoid repeating the phrase.
We can, for example, define a word to refer to a condition:
$$p_x := [2x = x^2] .$$

To state a problem is to state the condition
that its solutions have to satisfy.
This is the same as defining `something'
by listing the properties, uses, and, in general,
the requirements that anything has to fulfill
to be called `something'.
Perhaps because defining is like stating a problem,
it has not been studied in linguistics,
though it happens necessarily in a language.
But the first {\sc Lisp} of \cite McCarthy (1960)
already used definitions,
then introduced by the operation {\tt label},
``[i]n order to be able to write expressions
for recursive functions,'' page 186.
And that is because the genuine recursion
needs names to refer to functions.

{\sc Lisp} itself
would not loose any computing power
without {\tt label},
nor without its replacement {\tt define},
 see \cite McCarthy et al. (1962), \S2.3.
This is because {\sc Lisp} also implements
the functional semantics of lambda-calculus, and 
in lambda-calculus, which only uses anonymous functions,
a function can call itself using
the paradoxical combinator {\sf Y},
see \cite Curry \& Feys (1958), \S5G.
As \cite Friedman \& Felleisen (1987) show,
{\sc Lisp} does not need {\tt define},
but without names the code
goes easily beyond comprehension.

So, in the end, definitions are more than handy,
and we cannot go easily without them.
Our processing limitation to about seven chunks of information,
see \cite Miller (1956), can be the reason why
we real persons need defined words.

\subsection{Routine}

A \definition{resolution} is a way of solving a problem,
that is, a process that takes a problem
and tries to return the solutions to the problem.
A successful resolution of the problem
$x? [2x = x^2]$
can proceed this way:
$$\eqalign{
  [2x      &= x^2] \cr
  [2x - 2x &= x^2 - 2x] \cr
  [0       &= xx - 2x] \cr
  [0       &= (x - 2)x] \cr
  [x - 2 = 0]     &\lor [x = 0] \cr
  [x - 2 + 2 = 0 + 2] &\lor [x = 0] \cr
  [x = 2]         &\lor [x = 0] \cr
  \{2\} &\cup \{0\} = \{2,0\}\;.}$$
In this case the resolution was achieved by analogy
transforming the problem, or rather its condition,
until we found two subproblems with a known solution,
$x? [x=2]$ and $x? [x=0]$,
which we could then resolve by routine.
So the problem has two solutions, two and zero.

To represent each transformation the simplest expedient is
the ordered pair, $(s,t)$,
where $s$ and $t$ represent two expressions:
$s$ before being transformed, and
$t$ once it has been transformed.
Note that we can use a single ordered pair to
express each transformation, or the whole sequence.
For example, the complete sequence can be summarized
in just one ordered pair, which, in this case,
describes how to resolve the problem by routine,
because $s$ is the problem
and $t$ is the set of its solutions:
$$\left( x? [2x=x^2],\; \{0, 2\} \, \right) .$$

In summary:
To resolve a problem by routine you don't need to reason,
you only need to remember what are the solutions to the
problem, and that information can be expressed
using an ordered pair. Therefore,
we can use {\sc Lisp} to express any resolution by routine.
A resolution by routine can also be expressed
 using a two-element set,
though then we should mark at least one element in order to
distinguish which one is the problem and which one
the set of solutions, so Merge would also work.

\subsection{Trial}

We can also resolve by trial and error.
In the trial we have to test if a value satisfies the condition, or not,
and the condition is an open expression.
To note mathematically open expressions, also known as functions,
we will use lambda abstractions $(\lambda x\, p_x)$,
already seen in \ref{Recursive Language},
where $x$ is the free variable and $p_x$ the open expression.
In the case of our problem:
$$(\lambda x\, [2x=x^2]).$$

Now, to test a particular value $a$,
we have to bind that value $a$ to the free variable
inside the condition.
We also say that we apply value $a$ to the function $(\lambda x\, p_x)$.
In any case we write function applications this way: $((\lambda x\, p_x)\, a)$.
We can abbreviate the expression naming the function,
for example $f := (\lambda x\, p_x)$, to get
$(f\,a)$, which is just the lambda way to write the typical $f(a)$.
In our case, `to test if number $2$ is equal when it is doubled to
when it is squared' is written $((\lambda x\, [2x=x^2])\,2)$.
And to calculate if a value satisfies the condition,
we replace the free variable with the binding value;
this process is called $\beta$-reduction,
 also seen in \ref{Recursive Language}.
In our case we replace $x$ with $2$ ($x:=2$), this way:
$$((\lambda x\, [2x=x^2])\,2) \rightarrow [2.2 = 2^2] \rightarrow [4 = 4]
 \rightarrow \hbox{\sc true} .$$

Lambda abstraction, function application, $\beta$-reduction,
and an inexhaustible source of variables,
are all the ingredients needed to concoct
a lambda-calculus interpreter.
{\sc Lisp} implements all of them:
any non-reserved word can name a variable,
the reserved word {\tt lambda} as the left branch of a sentence
is a lambda abstraction,
and a lambda abstraction as the left branch of a sentence
is a function application on which a $\beta$-reduction
is automatically executed.
This way {\sc Lisp} implements the whole lambda functional semantics.

{\sc Lisp} implements the simplest $\beta$-reduction:
it uses a call-by-value reduction strategy, and
a positional assignment, that is,
the assignment of actual to formal parameters is based on position,
the first actual parameter to the first formal parameter,
the second to the second, and so on.
As with {\tt cons} and Merge, see \ref{Sentence},
natural language could be using
a more flexible kind of $\beta$-reduction,
as Unification proposed by \cite Jackendoff (2011).
Unification, see \cite Shieber (1986),
works on feature structures,
which are labeled tree data structures
that play the r\^ole of lambda abstractions,
and then any device implementing Unification
needs also operations to parse tree data structures,
as Merge. Therefore, Unification and Merge
can be complementary but not mutually exclusive.

A condition can only return two values,
which we will call {\sc true} and {\sc false}.
In case we want to make a difference depending on the condition,
and what else could we want?, then we have to follow
one path when the condition is satisfied ({\sc true}), and
a distinct path when the condition is not satisfied ({\sc false}).
{\sc Lisp} provides a conditional command {\tt cond}
that can do it, although it is more general:
$$\hbox{\tt (cond (} \left< \hbox{\it condition} \right>\;
\left< \hbox{\sc true \it case} \right> \hbox{\tt ) (t }
 \left< \hbox{\sc false \it case} \right>
 \hbox{\tt )} \hbox{\tt )}.$$

We can write down completely any trial
just by using these two expedients:
binding values to free variables in open expressions,
and a conditional command, as {\tt cond}.
Suppose, for example, that we guess that
the solution to our problem is one of the first four numbers,
in other words, that the solution is in set $\{1, 2, 3, 4\}$,
and that we want to try them in increasing order.
Then, defining $$f := (\lambda x\, [2x=x^2])$$
 and mixing notations freely, the trial would be:
$$\vbox{\tt \cleartabs
\+(cond &($(f\;1)$ 1)\cr
\+      &($(f\;2)$ 2)\cr
\+      &($(f\;3)$ 3)\cr
\+      &($(f\;4)$ &4)\cr
\+      &(t &nil)) {\rm.}\cr
}$$
In our natural language this code reads:
if the condition $f$ is satisfied by 1, then {\tt 1} is the solution;
or else if the condition $f$ is satisfied by 2,
 then {\tt 2} is the solution; and so on.
A conditional is the natural way to
express any trial and error resolution.

In fact, to code any computable function in {\sc Lisp},
that is, to code any Turing machine in {\sc Lisp},
we need the commands {\tt cond} and {\tt set!},
the predicate {\tt eq?}, and the procedures to control the tape.
From the inexhaustible source of words,
a finite set is needed to name the internal states, and
another finite set is needed to name the symbols on the tape.
Command {\tt set!}\ assigns a value to a variable, and
it is needed to set the internal state of the Turing machine.
Predicate {\tt eq?}\ compares two words, and it is needed
to determine which clause of the {\tt cond}-itional applies,
that is, which is the right clause for 
the current internal state and read symbol.
The procedures to control the tape
 of a Turing machine are just:
{\tt read}, which returns the word
 that is in the current position of the tape;
{\tt write}, which writes its argument
 in the current position of the tape; and
{\tt move}, which accepts as argument one of
{\it left}, {\it right}, or {\it halt},
and acts accordingly.
Finally, the conditional has to be executed in a loop.
This way {\sc Lisp} also implements the whole
functional semantics of computing.
You can find an example in \cite Casares (T) \S7.1.

\subsection{Analogy}

If we know the solutions of a problem,
then we can resolve it by routine,
which can be written as an ordered pair,
 $( x? [2x=x^2],\; \{0, 2\} \, )$,
as we have already done in \ref{Routine}.
If we do not know the solutions, but we suspect
of some possible solutions, then
we can resolve it by trial and error,
which can be written as a conditional,
$$\hbox{\tt (cond ($(f\;1)$ 1) ($(f\;2)$ 2)
 ($(f\;3)$ 3) ($(f\;4)$ 4) (t nil)){\rm,}}$$
as we have already done in \ref{Trial}.
There is a third way of solving problems: by analogy.

By analogy we transform a problem into another problem, or problems.
Most times the outcome will be more than one problem,
because `divide and conquer' is usually a good strategy for complex
problems. So, in general, the resolution of a problem will be a tree,
being the original problem its trunk. If we use analogy
to resolve it and we get, for example, four easier subproblems,
then the tree has four main branches. But, again, from
each branch we can use analogy, and then we get some sub-branches,
or we can use routine or trial. We resolve by routine
when we know the solutions, so the subproblem is solved;
these are the leaves of the resolution tree. Trial \dots

You got lost? Don't worry,
even myself get lost in this tree of trees,
and to what follows the only important thing to keep in mind is
one easy and true conclusion:
expressions that represent resolutions of problems
have a tree structure, because they describe the resolution tree.
And now we can answer the question that we left pending
at the end of Subsection \ref{Sentence}:
problem solving does indeed require a tree sentence structure.

Both {\sc Lisp} and natural language use
tree sentence structures, conditionals and pairs,
so both can be used to represent resolutions of problems.
The question now is to ascertain the additional
requirements that calculating resolutions of problems impose.
In particular, we will investigate the requirements that
a full problem solver has to fulfill in order to solve
as many problems as possible.

\subsection{Resolution}

A condition, also known as predicate, is a function
with two possible outcomes,
the loved {\sc true} and the hated {\sc false}.
This means that each time we apply it
we get a {\sc true} or a {\sc false}.
For example, defining again $f := (\lambda x\, [2x=x^2])$, we get:
$$\eqalign{
 (f\;0) &\rightarrow \hbox{\sc true}\cr
 (f\;1) &\rightarrow \hbox{\sc false}\cr
 (f\;2) &\rightarrow \hbox{\sc true}\cr
 (f\;3) &\rightarrow \hbox{\sc false}\cr
 (f\;4) &\rightarrow \hbox{\sc false}\cr
 \omit\hfil\vdots\hfil\span\omit\cr}
 $$

What solves definitively a problem is the inverse function
of the condition of the problem.
Because the inverse function just reverses the condition,
from a
 $0\rightarrow \hbox{\sc true}$
to a
 $\hbox{\sc true}\rightarrow 0$,
so when we apply {\sc true} to the inverse condition we get the solutions,
and when we apply {\sc false} we get the set of the no-solutions.
Thus we get:
$$\eqalign{
 (\check f\; \hbox{\sc true}) &\rightarrow \{0, 2\}\cr
 (\check f\; \hbox{\sc false}) &\rightarrow \{1,3,4,\dots\} .\cr
 }$$

We can use the condition in both directions:
the natural direction,
 when we apply a value to test if it satisfies the condition, and
the opposite direction,
 when we want to know what values satisfy the condition.
To express a problem is enough to write the condition and
to indicate which are its free variables, and
to solve a problem is enough to apply the condition in the
opposite direction.

It is too easy to say that the inverse function of its condition
solves the problem; I have just done it.
Unfortunately, it is nearly always very difficult to find
the inverse of a condition, and sometimes the condition is
inexpressible or unknown.

We should distinguish solving,
which is finding solutions, as the inverse condition does,
from resolving, which is finding resolutions, that is,
finding ways of finding solutions. Then
resolving is calculating the inverse function of a condition,
given the condition, and therefore a resolution is a function that,
when it is successful, takes a problem, or its condition,
and returns the problem solutions.
$$\hbox{Problem}
\mathop{\hbox to 80pt{\rightarrowfill}}\limits^{\hbox{\rm Resolution}}
\hbox{Solution}$$

In any case, finding a resolution to a given problem is a problem,
its \definition{metaproblem}.
Finding a resolution to a problem is a problem because
it has its two mandatory ingredients.
There is freedom, because there are several ways to solve a problem,
which are the routine and the different trials and analogies, 
and there is a condition, because not every function
is a valid resolution for a problem,
but only those that return the solutions to the problem.
And then, being a problem, we need a resolution
to find a resolution to the problem. And, of course,
to find a resolution to find a resolution to the problem we need,
can you guess it?, another resolution.
What do you think? Better than `convoluted' say `recursive'.

\subsection{Recursivity}

The solution to a problem can also be a problem or a resolution.
For example,
when a teacher is looking for a question to ask in an exam,
her solution is a problem.
And when an engineer is designing an algorithm
to calculate some type of electrical circuits,
her solution is a resolution.
In this last case, the next problem could be
to enhance the algorithm found,
and then the resolution of this optimization problem
would take a resolution to return a better resolution.
Solutions can also be enhanced.

The condition of a problem takes a possible solution as argument
and returns {\sc true} when the argument is a solution,
and {\sc false} when it is not a solution. Therefore,
if the solutions of a problem are problems, then
the condition of the problem has to take problems,
 or conditions, as arguments, and
if the solutions of a problem are resolutions, then
the condition of the problem has to take resolutions as arguments.
This last case is the case of metaproblems, for example.

A full problem solver has to solve as many problems as possible,
and then it should take any problem, or its condition,
and it should return the problem solutions.
Any condition is a problem,
if we are interested in the values that satisfy the condition,
and then the full problem solver has to take any condition as input,
including conditions on problems and on resolutions.
Not every function is a condition,
but only generalizing to functions
we can cope with the resolution to resolution
transformation that the engineer above needed.
And the solutions returned by the full problem solver
can be also problems and resolutions.

The conclusion is then that a full problem solver resolution
should input any expression and can output any expression.
So a full resolution has to be able to transform
any expression into any expression.
$$\left.\vcenter{\halign{\hfil\rm#\crcr
 Problem\cr Resolution\cr Solution\cr}}\right\}
 \mathop{\hbox to 80pt{\rightarrowfill}}\limits^{\hbox{\rm Resolution}}
 \left\{\vcenter{\halign{\rm#\hfil\crcr
 Problem\cr Resolution\cr Solution\cr}}\right.
$$
The resolution can then be the transformation,
but also what is to be transformed,
and what has been transformed into.
We call this property of transformations
that can act on or result in transformations
without any restriction \definition{recursivity}.

We can now answer the question
asked at the end of Subsection \ref{Analogy}.
Full problem solving requires recursion,
because full problem solving requires
functions taking functions and returning functions
as generally as possible, without restrictions,
and then a whole functional semantics is needed.
The function is the mathematical concept for transformation,
for change. Therefore,  provided with functional semantics,
the recursive problem solver masters change.

When the current state of things makes survival difficult
or inconvenient or simply improvable,
the problem is to find a way to go to a better state.
In these circumstances,
a living being that does not implement recursion
can try some possible ways of solving from a set engineered
by evolution, which can include
imitating others, or
trying with sticks or some other tools.
But a recursive individual, with functional semantics,
can calculate any change and its results
before executing one.
Being a tool any means to make change easier,
the calculation can introduce a subproblem that asks for a tool,
and then the Turing complete individual can devise the proper tool,
and even devise a tool to make the tool, and so on. 
Mastering change, we recursive solvers can foresee
the effects of our actions,
and, at least to some extent,
we can match the future to our own will.

\subsection{Quotation}

Recursivity needs a quoting expedient to
indicate if an expression is referring to a transformation,
or it is just an expression to be transformed.

Joke:\\
\null\qquad --- Tell me your name.\\
\null\qquad --- Your name.\\
Who is answering has understood `tell me ``your name''\thinspace',
which could be the case, although unlikely.
In any case, quotation explains the confusion.

\goodbreak
\setbox2=\hbox{$\vdots$}\ht2=0pt\dp2=0pt

Quotation is very powerful.
Suppose that I write:
\begingroup\everypar{}
\medskip
\vbox{\advance\leftskip20pt\advance\rightskip20pt
  \abovedisplayskip=6pt\belowdisplayskip=6pt\sl
  En cuanto llegu\'e, el ingl\'es me dijo:
  ``I was waiting you to tell you something. \dots
 $$\matrix{\copy2\cr
   \hbox{\rm[a three hundred pages story in English]}\cr
   \raise-4pt\box2\cr}$$
 \dots And even if you don't believe me, it is true''.
 Y se fue, sin dejarme decirle ni una sola palabra.}
\smallskip
Technically, I would have written a book in Spanish!
\endgroup

Technically, using the quoting mechanism,
all recursive languages are one.
This is because, by using quotation,
recursive languages can incorporate
 any foreign or artificial construction.
For example, in a {\sc Lisp} manual written in English,
 as \cite McCarthy et al. (1962),
there would be some parts following the English syntax,
and some other quoted parts following the {\sc Lisp} syntax
that is being explained in the manual.
{\sc Lisp} is a recursive language because
a {\sc Lisp} interpreter can calculate any recursive function,
and then English is a recursive language because
we can learn {\sc Lisp} from a manual in English.

That all recursive languages are one
is in line with Chomsky's universal grammar,
and with his idea that seen globally
``there is a single human language,
with differences only at the margins'',
quoted from \cite Chomsky (2000), page 7.

Mathematically,
we have already dealt with this situation
 in \ref{Universal Grammar Is Universal}:
Any recursive language can produce any recursive language,
and even any language,
because the device that implements the recursive language
is a universal computer, that is, a universal grammar.

\subsection{Lisp}

Lastly, we should review a complete
{\sc Lisp} interpreter.
Instead of the first {\sc Lisp} proposal
by \cite McCarthy (1960), we will analyze
the complete implementation
by \cite Abelson \& Sussman (1985).
In fact, they implement two {\sc Lisp} interpreters:
one coded in {\sc Lisp} in \S4.1, and
another one coded on a register machine in \S5.2.
Both interpreters code a read-eval-print loop,
 \S4.1.4 and \S5.2.4,
where the core is the evaluator,
 \S4.1.1 and \S5.2.1.
The evaluator is a dispatcher.

The first task of the evaluator is to treat the
self-evaluating words.
Self-evaluating words have a meaning by themselves,
and therefore they do not need any syntactic treatment.
So this is the interface to semantics.
In the case of this {\sc Lisp} interpreter by Abelson \& Sussman,
only numbers are self-evaluating,
but on commercial ones there can be several
kinds of semantic words,
as booleans or strings or vectors.
In the case of natural language,
there are thousands of self-evaluating words,
also known as content or autosemantic words,
mainly nouns, adjectives, verbs, and adverbs.

The second task is to deal with quoted text,
which is also excepted of syntactic treatment.
This {\sc Lisp} interpreter implements only {\tt quote},
which is the literal quotation operation, but
commercial ones have additional quotation operators,
as {\tt quasiquote} or {\tt macro}.
Using a general version of quotation,
like the one used in written natural language,
each kind of quotation would be dispatched
to its proper parser, and then a book in English
should quote differently its parts in Spanish
and its parts in {\sc Lisp}, for example.

After the exceptions, the dictionaries, where
a dictionary is a list of variable-value pairs,
and a variable can be any non-reserved word.
This {\sc Lisp} has three kinds of variables:
\begin{itemize}
\item from genuine recursion,
those introduced by {\tt define}, which cannot be modified,
and then they are like technical definitions in natural language;
\item from computing,
those introduced by {\tt set!}, which can be modified
and have some similarities to the pronouns of natural language; and
\item from lambda-calculus,
those introduced by lambda abstractions,
which are the names of the function arguments,
and then, comparing a function to a verb,
these variables are the subject and objects of the sentence,
or clause.
\end{itemize}
This {\sc Lisp} is strict and every variable has a well-defined
scope, so its interpreters have to construct
a hierarchy of dictionaries, and their look-up procedures
have to start from the nearest one at the bottom,
and they have to go up to the next one when searching fails.

Then the evaluator deals with {\tt lambda} and {\tt cond}.
The first implements the lambda abstraction of lambda-calculus,
and the second the conditional of computing.
These two plus the {\tt define} of genuine recursion
are each one the core of a functional semantics.
Natural language can be using a more flexible
version of lambda abstraction, as the
feature structure of Unification, for example,
that could work also as a dictionary.

The only remaining task of the evaluator is
to apply functions.
This {\sc Lisp} has two kinds of functions:
primitive functions, which are implemented by the interpreter,
and compound functions, which are calculated by $\beta$-reduction.
This {\sc Lisp} interpreter implements
the following primitive functions:
\begin{itemize}
\item {\tt cons}, {\tt car}, {\tt cdr}, and {\tt atom?},
which are needed to implement its syntax;
\item {\tt eq?}, which is needed to implement
the functional semantics of computing;
\item some procedures needed to access media,
as {\tt read}, {\tt write}, and {\tt move}, for tapes;
\item some procedures needed to deal with
self-evaluating, or semantic, words; and
\item some other functions implemented just for efficiency.
\end{itemize}
Because numbers are self-evaluating for this interpreter,
then it has to implement some primitive functions for numbers,
as the successor function, {\tt 1+},
or the predicate {\tt number?}, for example.
In the case of natural language,
with thousands of semantic words,
we should expect to find also thousands of primitive functions,
and though most of them would not be language specific,
as for example the one for face recognition,
some would be, as Merge for syntax and also
those needed to access media providing an interface to
the sound system for speech generation and recognition.
And, instead of $\beta$-reduction,
natural language can be using a more flexible
version, as Unification, for example.

Comparing {\tt cons} and its sibling syntax operations
with the complete {\sc Lisp} interpreter,
the conclusion should be again the same:
the very reason of any recursive device is not syntax parsing,
but full problem solving.
Then, the interface to pragmatics is
the very reason of our natural recursive machinery.

\needspace{30pt}

{\it The devil is in the details}.
To study the evolution of the eye,
it is important to know the laws of optics.
A video camera does not work as an eye
because the materials and methods of an engineer are
different from those of evolution,
but the limitations of optics are the same for both.
Here, the {\sc Lisp} interpreter plays the r\^ole
of the video camera, providing
a complete and detailed list with all the requirements
 that a recursive device has to implement, and
a working model with the reasons why
 each and every element of the whole 
 does what it does.
Unfortunately, we lack other living implementations
to compare human recursion with,
but at least we have {\sc Lisp}.
With this warning note about the use of models,
we conclude our tour around {\sc Lisp},
 a masterpiece of engineering.

\subsection{Summary}

The requirements found
 in this Section~\ref{Evolution}
are evolutionary,
so they are not pre-conditions of design,
but post-conditions.
Each evolutionary requirement appears by chance,
grows in the short term by a mixture of chance and fitness,
is kept in the mid term because of its fitness,
and is entrenched in the long term as part of a bigger structure,
which is the recursive engine, in our case.

From a mathematical point of view,
the list of requirements found
 in this Section~\ref{Evolution} is redundant.
For example, lambda-calculus does not need definitions,
because it does everything with anonymous functions.
But, for a function to refer to itself, we need names, and
therefore to define genuine recursive functions we need definitions!
It is nearly the same with conditionals.
Lambda-calculus does not need them because they can
be defined as any other function,
but the conditional is an instruction that
a universal Turing machine needs.
That the list of requirements is redundant only means
that evolution does not value succinctness, but efficiency.
And though the different versions of recursion are
mathematically equivalent, their implementations are
distinct from an engineering point of view.
Then, by implementing more than one,
the problem solver has the opportunity to use each time
the version that is best suited to the problem.
And redundancy also pays in robustness.

From a very abstract point of view,
all the requirements found in this Section~\ref{Evolution}
can be connected to the concept of function,
 which is the key concept of recursion.
A function is the general form of conditional,
because the result of the function is conditioned
 on the values it takes;
a computable function calculated by a universal Turing machine
is expressed just as a list of conditionals.
But a function can also be expressed as a structure with
 holes that, depending on how the holes are filled,
results differently.
The holes are pronouns, and the rest of the structure
are other words that compose the sentence.
The pair is arguably the simplest structure,
and then a tree built from pairs is surely
the simplest open structure, and therefore
the binary tree is the first candidate for the sentence.
See that the Merge operation of the minimalist program
builds binary trees.
And, finally, we need recursion to get the
most general functions that are still effectively calculable,
as first stated by \cite Church (1935).

The point of this Section~\ref{Evolution}
was to show some requirements from full problem solving
that our natural languages exhibit.
This does not prove that they were originated by
problem solving,
 and below in Subsection~\ref{Syntax and Evolution}
 we will argue that some were possibly not,
but, in any case,
they enhanced our problem solving abilities, and then
these `problematic' features of language
enhanced our survival fitness because of problem solving.

\section{Discussion}

\subsection{Problems and Evolution}

Life can be assimilated to the survival problem.
From this point of view,
which we will call the problematic view of life,
evolution is a problem solver.
Evolution solved some subproblems
of the survival problem by designing systems,
as the cardiovascular system, and
it solved their sub-subproblems by designing organs,
as the heart.

But evolution cannot solve the problems faced
by individuals in their day-to-day living,
because those problems depend on casual circumstances.
For solving those problems faced by individuals,
evolution designed the nervous system, the brain organ,
and even the specialized nervous tissue and neuron cell.
Then, broadly speaking, the function of the nervous
system is to deal with information, and the
function of the brain is 
to take information from the body,
compute what to do in order to resolve
 according to the circumstances, and
send the resulting command information back to the body.
In other words, the brain is the solver
of the problems of the individual.

There is a delicate interaction between the brain
and the rest of the body, which is calibrated by the
proper distribution of responsibilities between
the two problem solvers involved, evolution and the brain.
For example, heart beat rate can be autonomous,
as shown by a separated heart beating,
but the brain can command to increase the beat rate
when running, for example, to flee from a predator.

We are Turing complete because our brain is Turing complete,
but not all living individuals are Turing complete,
so we might wonder what difference does this make.
To start, we have to distinguish
solving, or finding solutions, from
resolving, or finding resolutions, where
a resolution is a way of solving, 
that is, a process for solving problems,
and then mathematically, as seen in \ref{Resolution},
a resolution is a function that takes a problem
and returns solutions, right and wrong solutions.
So, being a function, a Turing complete individual
can express and understand, that is, imagine,
any resolution in her recursive language,
while  a more limited individual will apply
its limited set of resolutions to any problem.
The key point is that
a single Turing complete individual
can imagine any possible resolution,
that is, any possible way of solving a problem,
and then she can execute any of the imagined resolutions
 that returns right solutions,
while an individual that is not Turing complete
can only apply those resolutions that are
implemented in the hardware of its body,
mainly in the hardware of its brain.

Species that are not Turing complete
need a genetic change to modify their set of resolutions,
while Turing complete individuals
can apply new resolutions without any hardware change,
but just by a software change.
So the timing of creativity depends on
evolution until Turing completeness is achieved,
and it does not depend on evolution after that point
for the species that achieve Turing completeness.
We will call every point where an evolutionary path
achieves Turing completeness an
\definition{evolutionary singularity}.
In other words, an evolutionary singularity is
any evolutionary moment when the brain surpasses evolution
in problem solving.

New behaviors related to the solution of problems,
such as feeding, mating, and, in general, surviving,
should proliferate after an evolutionary singularity.
And noting that
a tool is the physical realization of a resolution,
then an explosion of new tools should start
whenever an evolutionary singularity happens.
So the archaeological record of human tools
should point to our evolutionary singularity.

In summary,
creativity is slow until an evolutionary singularity
and creativity explodes after every evolutionary singularity
because,
after achieving Turing completeness,
performing new ways of solving the survival problems
becomes cheap and available to single individuals
while, before achieving it,
any single new way of solving them required
genetic changes on species over evolutionary time spams.

Nevertheless, Turing completeness is defined by
a single condition: pan-computability.
A universal Turing machine,
which is the prototype of Turing completeness,
is every Turing machine that can compute
whatever any Turing machine can compute, but no more.
This means that to perform any specific computation you can,
either build the Turing machine specific for that computation,
or write the program for that specific computation
 on a universal Turing machine.
Either way the outcome of the computation will be the same,
and the only differences would be that
the first can run faster, and that
the second can be implemented faster,
 once you have a universal computer.
The second is better for modeling, that is, for imagining,
because writing models is much faster than building models,
but, again, the results of the computations are the same.

In other words, creativity is the only exclusive feature
of Turing complete problem solvers.
And this explains an elusive fact:
every time a specific behavior is presented as uniquely human,
it is later rejected when it is found in another species.
The point is not that
we behave in some specific way to solve a problem,
but that
we are free to imagine any way to solve our problems.
Creativity is the mark of Turing complete solvers.

\subsection{Syntax and Problems}

We could equate problem solving with syntax, because
syntax is needed just to express problems,
but mainly because
both syntax and problem solving are computing.
And full problem solving requires universal computability,
which is the maximum syntactic requirement.
So full problem solving needs syntax, and needs all of it:
full problem solving needs a complete syntax.
$$\left.\matrix{\hbox{Complete}\cr\hfill\hbox{Syntax}\cr}\right\} = 
 \left\{\matrix{\hbox{Universal}\cr\hbox{Computing}}\right\} =
 \left\{\matrix{\hbox{Full}\hfill\cr\hbox{Problem Solving}}\right.$$

By identifying syntax with problem solving
we have solved the syntax purpose paradox
presented in \ref{Chomsky's Identity}:
when a brain becomes a complete syntax engine,
then that brain becomes a full resolution machine.
So syntax can be meaningless, but it is very useful
and its purpose is to solve problems in any possible way.
The key is that while a complete syntax is still mechanical,
it is not really meaningless,
because it has a functional semantics.

That both syntax and problem solving are computing
is a coincidence that
can also be derived from the problematic view of life.
In this view, whatever the brain does is to compute
for solving the problems of the individual.
Communication, or sharing information with other individuals,
is just one problem out of those that
the brain has to solve, and
language is just a particular solution for communication.
This explains the coincidence, but here
we distinguish communication from thinking,
defining very artificially thinking as what the brain does
that is not communication, to separate the linguistic requirements
from the much more general cognitive requirements.
The reason is that syntax,
which is in principle a solution to the communication problem,
was instrumental in achieving full problem solving,
which concerns cognition generally.

Syntax and problem solving use the same computing machinery,
but their particular requirements are different:
better syntax is for better communication, while
better problem solving is for better thinking.
Then our computing machinery should have evolved
by satisfying requirements
from syntax (communication), or
from problem solving (thinking), or
from both. And then we should find
problem solving requirements in syntax, and
syntax requirements in problem solving,
or, in fewer words,
language and thinking co-evolved in humans.

The description by \cite Vygotsky (1934) of the
co-development of language and thinking in children
can be seen as a possible model for the
co-evolution of language and thinking under our hypothesis.
In fact, at this very abstract level,
the development of syntax and problem solving in children
can use every argument used here to base our hypothesis,
and then, the very same arguments can be used
to base a parallel hypothesis:
`syntax and problem solving co-develop in children
 towards recursion'.
This parallel hypothesis is easier to examine,
and if it is found true, then the original hypothesis,
which is based on the same arguments
though applied to a different process,
would gain some support.

Syntax recursivity is credited with
making possible the infinite use of finite means,
which is obviously true.
But our wanderings in the problem resolution realm
have shown us that there are other features
provided by recursion that are,
at least, as important as infinity;
infinity that, in any case, we cannot reach.
The main three ones are:
 sentences, or hierarchical tree structures;
 functions, or open expressions with free variables; and
 conditionals, or expressing possibilities:
{\it I cannot imagine how I would see the world
if there were no conditionals, can you?}
Conditionals, for example,
are necessary to think plans in our head
that can solve our problems, and then
problem solving gives conditionals a purpose
and an evolutionary reason that they cannot get
from language.

\subsection{Syntax and Evolution}

We have seen that syntax and problem solving
 co-evolved towards Turing completeness,
because both are computing, and
 Turing completeness is the maximum computing power.
But, how was that co-evolution?,
that is, which one,
syntax or problem solving, was driving
each step of the co-evolution?

Before answering these questions,
we must set the starting point.
As syntax requires words to work on,
we will start when there were already words.
\cite Pavlov (1927) showed that
the assignment of arbitrary sounds to meanings
is a capability that dogs already have,
and then words should have preceded syntax.
But dog words are indices for \cite Pierce (1867),
and neither icons nor indices can be produced
in high quantities, so we should expect to
have symbols before syntax developed.
For these and other reasons,
our starting point will be a protolanguage,
as presented by \cite Bickerton (1990),
that has already developed phonology and
a lexicon with its semantics.

We argue that the very first steps,
leading to the sentence, were driven by syntax.
The main reason is that syntactic languages are
exponentially more expressive than asyntactic languages,
see \cite Casares (H).
By giving meaning to sentences, a syntactic language
can assign different meanings to word sequences as
`dog bit child' and `child bit dog' that would mean
exactly the same in asyntactic languages.
The implication is that ambiguity is reduced efficiently
just by imposing some structure to a set of words,
and this is the case even without free variables,
which was possibly the first reason
why problem solving needed sentences.
For example,
the first step of syntax could be very simple:
just put the agent before every other word.
This simple expedient by itself would
prevent efficiently some ambiguities,
explaining who did and who was done something.
Marking who is the agent would also achieve it.
And that natural languages do not provide
inexhaustible sets of variables
points in this direction, too.

We argue that the last steps,
fulfilling universal computability,
were driven by problem solving.
The reason, now, is that Turing completeness
is useful for problem solving to
detach itself from hardware causing an explosion of creativity,
but it is detrimental to natural language syntax,
as stated by the anomaly of syntax.

I don't know which one, syntax or problem solving,
was driving each of the other steps that provided
each of the capabilities that are
needed to implement a Turing complete brain.
But what I take for sure is that
we humans evolved to Turing completeness,
not to achieve a full resolution machinery,
which was the final prize,
but because each recursive feature,
such as the sentence, the open expression, and the conditional,
made thinking (problem solving) or communication (syntax), or both,
more efficient, resulting in more fitted individuals.
A more detailed study of the requirements
found in Section~\ref{Evolution},
in addition to those found in a grammar text book,
will be needed to disentangle the co-evolution
of syntax and problem solving.

Our general view of the evolution
from an animal communication system to
our recursive language has then a middle point,
protolanguage.
$$\hbox{Animal Communication}\longrightarrow
 \hbox{Protolanguage}\longrightarrow
 \hbox{Recursive Language}$$
To reach the protolanguage,
evolution had to provide facilities that were both
cognitive, to store and retrieve many words, and
anatomical, to utter and decode many words.
Tree data structures are simple and useful,
so they were probably already used for vision,
see \cite Marr (1982),
and then the next step was to repurpose,
or just to reinvent,
the tree data structure computing capability
to use it with words, yielding syntax.
After syntax,
some more evolutionary steps were required to
implement a functional semantics,
reaching Turing completeness.
Therefore,
the evolution from protolanguage to recursion
was only cognitive, or more precisely,
it was completely computational, and then
it should have left no anatomical clues,
and it could have happened entirely
within the time frame of 
the anatomically modern {\it Homo sapiens}.

In any case, it would be difficult to deny that
syntax was, at least,
instrumental in achieving Turing completeness,
and therefore that syntax was influential
in reaching our evolutionary singularity.

\subsection{Beyond the Singularity}

These results should help us to reevaluate syntax.
In language evolution there are two main positions
regarding syntax; see \cite Kenneally (2007), Chapter 15.
The gradualist side defends a more gradual evolution,
where syntax is just a little step forward
that prevents some ambiguities and
makes a more fluid communication;
 see \cite Bickerton (2009), page 234.
For the other side, led by Chomsky, syntax is
a hiatus that separates our own species from the rest;
see \cite Hauser, Chomsky, and Fitch (2002).
What position should we take?

The co-evolution of syntax and problem solving explains that,
once our ancestors reached Turing completeness,
they acquired complete syntax,
also known as recursion, and thus they mastered change
and they became full problem solvers.
Then they could see a different the world.
How much different? Very much.

Seeing the world as a problem to solve
implies that we look for the causes that are hidden
behind the apparent phenomena.
But it is more than that.
Being able to calculate resolutions inside the brain
is thinking about
plans, goals, purposes, intentions, doubts, possibilities.
This is more than foreseeing the future,
it is building the future to our own will.
Talking about building, what about tools?
A tool is the physical realization of a resolution,
and then with complete syntax we can design tools,
and even design tools to design tools,
and so on recursively.

We are creative because we master change.
With our complete syntax,
we can imagine any possible transformation and its effects,
so we can create our own imagined worlds,
and we can drive the real world in the direction we want,
creating artificial domains in it.
Mainly, we adapt the world to us, instead of the converse.
Rather than just surviving,
we are always looking for new ways to improve our lives.

We need syntax to express the freedom of problems,
and with complete syntax we can
calculate any way of solving problems. Note that
calculating different ways of solving our problems is 
calculating how to use our freedom in order to achieve our ends.
So, yes, we are free because of syntax.

Summarizing,
the identification of syntax with problem solving
explains why syntax, being so little thing,
has made us so different.
Then we agree with Chomsky on that
syntax was instrumental in creating the hiatus
that separates our species from the rest.
But recursion is more than just one operation,
so we have probably acquired the complete syntax
in more than one evolutionary step,
and on this we agree with the gradualist party.

\section{Conclusion}

Only our species is Turing complete. Therefore,
we must explain the evolution of Turing completeness
to understand our uniqueness.

Turing complete individuals can transform strings of symbols,
irrespective of the symbols meanings,
but according to any possible finite set of well-defined rules.
It seems a nice capability but, being meaningless, not very useful.
It could be, but syntax is also about
meaningless transformations of strings of symbols according to rules.
Turing completeness is then a pan-syntactical capability,
because the syntax of a natural language does not need to follow
any possible set of rules, but only one specific set of rules.

Syntax is very peculiar indeed, because
syntax is a peculiarity of human language,
which is our most peculiar faculty.
For this reason,
we can argue that syntax is what defines our species,
and yet it seems improbable to explain how
some mechanical transformations of strings of symbols
have made us like we are.
In addition, syntax is not our only peculiarity:
is it a coincidence that we are the only species
with a syntactic language and also the most creative?

But then, after realizing that
Turing completeness is closely related to syntax,
which is a human peculiarity, we have to halt,
because we cannot progress anymore without new inputs.
In order to overcome that impasse,
this paper introduces a new piece of the jigsaw puzzle
that links areas that were previously unconnected:
the new piece is problem solving.
Problem solving is a piece of problem that
fits with syntax on one side,
and with Turing completeness on the other side.

We are now ready to understand Turing completeness
from the perspective of problem solving.
After finding that creativity is the peculiarity
of Turing complete problem solvers, we see
how our Turing complete problem solving machinery explains
both our creativity and our pan-syntactical capability.
But before that, we have also seen that
Turing complete problem solving needs
sentences, functions, and conditionals;
all of them employed by syntax.

The conclusion is that
syntax and problem solving should have
co-evolved in humans towards Turing completeness.

\xsection{Acknowledgements}

Thanks to
Kleanthes Grohmann for his encouragement,
and to
Mark Hauser,
Ray Jackendoff,
Diego Krivochen,
and the anonymous reviewers of Biolinguistics,
for their comments
on previous versions of this paper.

\xsection{References}
\everypar{} \vskip\parskip \parskip=0pt \frenchspacing

\reference Abelson \& Sussman (1985):
Harold Abelson, and Gerald Sussman, with Julie Sussman,
\book{Structure and In\-terpretation of Computer Programs};
The MIT Press, Cambridge MA, 1985,
\ISBN 978-0-262-01077-1.

\reference Bickerton (1990):
Derek Bickerton,
\book{Language \& Species};
The University of Chicago Press, Chicago, 1990,
\ISBN 978-0-226-04611-2.

\reference Bickerton (2009):
Derek Bickerton,
\book{Adam's Tongue: How Humans Made Language, How Language Made Humans};
Hill and Wang, New York, 2009,
\ISBN 978-0-8090-1647-1.

\reference Bickerton (2014):
Derek Bickerton,
\book{More than Nature Needs: Language, Mind, and Evolution};
Harvard University Press, Cambridge MA, 2014,
\ISBN 978-0-674-72490-7.

\reference Casares (H):
Ram\'on Casares,
``A Complete Hierarchy of Languages'';\\
\DOI{10.6084/m9.figshare.6126917}.

\reference Casares (P):
Ram\'on Casares,
``Problem Theory'';\\
\DOI{10.6084/m9.figshare.4956353},
\URL{\tt arXiv:1412.1044}<http://arxiv.org/abs/1412.1044>.

\reference Casares (R):
Ram\'on Casares,
``On `On Recursion'\thinspace'';\\
\DOI{10.6084/m9.figshare.5097691}.

\reference Casares (T):
Ram\'on Casares,
``On Turing Completeness, or Why We Are So Many'';\\
\DOI{10.6084/m9.figshare.5631922}.

\reference Casares (U):
Ram\'on Casares,
``Universal Grammar is a universal grammar'';\\
\DOI{10.6084/m9.figshare.4956764}.

\reference Chomsky (1959):
Noam Chomsky,
``On Certain Formal Properties of Grammars'';
in \periodical{Information and Control},
vol.\ 2, no.\ 2, pp.\ 137--167, June 1959,
\DOI{10.1016/S0019-9958(59)90362-6}.

\reference Chomsky (1965):
Noam Chomsky,
\book{Aspects of the Theory of Syntax};
The MIT Press, Cambridge MA, 1965,
\ISBN 978-0-262-53007-1.

\reference Chomsky (2000):
Noam Chomsky,
\book{New Horizons in the Study of Language and Mind};
Cambridge University Press, Cambridge, 2000,
\ISBN 978-0-521-65822-5.

\reference Chomsky (2005a):
Noam Chomsky,
``Some Simple Evo-Devo Theses:
 How True Might They Be for Language?'';
Paper of the talk given to the 
Morris Symposium on the Evolution of Language,
held at Stony Brook University, New York,
in October 15, 2005;
{\sc url:}
\URL<https://linguistics.stonybrook.edu/events/morris/05/program>.

\reference Chomsky (2005b):
Noam Chomsky,
``Three Factors in Language Design'';
in \periodical{Linguistic Inquiry},
vol.\ 36, no.\ 1, pp.\ 1--22, Winter 2005,
\DOI{10.1162/0024389052993655}.

\reference Chomsky (2006):
Noam Chomsky,
\book{Language and Mind}, Third Edition;
Cambridge University Press, Cambridge, 2006,
\ISBN 978-0-521-67493-5.

\reference Chomsky (2007):
Noam Chomsky,
``Of Minds and Language'';
in \periodical{Biolinguistics},
vol.\ 1, pp.\ 9--27, 2007,\\
{\sc url:}
\URL<http://www.biolinguistics.eu/index.php/biolinguistics/article/view/19>.


\reference Church (1935):
Alonzo Church,
``An Unsolvable Problem of Elementary Number Theory'';
in \periodical{American Journal of Mathematics},
vol.\ 58, no.\ 2, pp.\ 345--363, April 1936,
\DOI{10.2307/2371045}.
Presented to the American Mathematical Society,
April 19, 1935.

\reference Curry \& Feys (1958):
Haskell B.\ Curry, and
Robert Feys, with
William  Craig,
\book{Combinatory Logic}, Vol.\ I;
North-Holland, Amsterdam, 1958,
\ISBN 978-0-7204-2207-8.

\reference Davis (1965):
Martin Davis (editor),
\book{The Undecidable: Basic Papers on Undecidable Propositions,
Unsolvable Problems and Computable Functions};
Dover, Mineola, New York, 2004,
\ISBN 978-0-486-43228-1.
Corrected republication of the same title
by Raven, Hewlett, New York, 1965.

\reference Davis (1982):
Martin Davis,
``Why G\"odel Didn't Have Church's Thesis'';
in \periodical{Information and Control},
vol.\ 54, pp.\ 3--24, 1982,
\DOI{10.1016/s0019-9958(82)91226-8}.

\reference Everett (2008):
Daniel L.\ Everett,
\book{Don't Sleep, There Are Snakes:
Life and Language in the Amazonian Jungle};
Vintage, New York, 2008,
\ISBN 978-0-307-38612-0.

\reference  Fitch, Hauser, and Chomsky (2005):
Tecumseh Fitch, Marc Hauser, and Noam Chomsky,
``The Evolution of the Language Faculty:
Clarifications and Implications'';
in \periodical{Cognition},
vol.\ 97, pp.\ 179--210, 2005,
\DOI{10.1016/j.cognition.2005.02.005}.

\reference Friedman \& Felleisen (1987):
Daniel Friedman, and Matthias Felleisen,
\book{The Little LISPer}, Trade Edition;
The MIT Press, Cambridge, Massachusetts, 1987,
\ISBN 978-0-262-56038-2.

\reference G\"odel (1930):
Kurt G\"odel, 
``\"Uber formal unentscheidbare S\"atze der Principia Mathematica
 und verwandter Systeme I'';
in \periodical{Monatshefte f\"ur Mathematik und Physik},
vol.\ 38, pp.\ 173--198, 1931,
\DOI{10.1007/BF01700692}.
Received November 17, 1930.
Extended by \cite Rosser (1936).
English translation in \cite Davis (1965).

\reference Hamblin (1973):
Charles Hamblin,
``Questions in Montague English'';
in \periodical{Foundations of Language},
vol.\ 10, pp.\ 41--53, 1973.

\reference Hauser, Chomsky, and Fitch (2002):
Marc Hauser, Noam Chomsky, and Tecumseh Fitch,
``The Language Faculty:
 Who Has It, What Is It, and How Did It Evolved?'';
in \periodical{Science} 298, pp.\ 1569--1579, 2002,
\DOI{10.1126/science.298.5598.1569}.

\reference Jackendoff (2011):
Ray Jackendoff,
``What Is the Human Language Faculty?: Two Views'';
in \periodical{Language}, 
vol.\ 87, no.\ 3, pp.\ 586--624, September 2011,
\DOI{10.1353/lan.2011.0063}.

\reference Kenneally (2007):
Christine Kenneally,
\book{The First Word: The Search for the Origins of Language};
Penguin Books, New York, 2008,
\ISBN 978-0-14-311374-4.

\reference Kleene (1936a):
Stephen Kleene, 
``General Recursive Functions of Natural Numbers'';
in \periodical{Mathmatische Annalen},
vol.\ 112, pp.\ 727--742, 1936,
\DOI{10.1007/BF01565439}.

\reference Kleene (1936b):
Stephen Kleene,
``$\lambda$-Definability and Recursiveness''; 
in \periodical{Duke Mathematical Journal},
vol.\ 2, pp.\ 340--353, 1936,
\DOI{10.1215/s0012-7094-36-00227-2}.

\reference Kleene (1952):
Stephen Kleene,
\book{Introduction to Meta-Mathematics};
Ishi Press, New York, 2009,
\ISBN 978-0-923891-57-2.
Reprint of the same title by
North-Holland, Amsterdam, 1952.

\reference Krifka (2011):
Manfred Krifka,
``Questions'',
\DOI{10.1515/9783110255072.1742};\\
in \book{Semantics:
 An International Handbook of Natural Language Meaning},
 Volume 2, HSK 33.2,
 K.\ von Heusinger, C.\ Maienborn, P.\ Portner (editors),
pp.\ 1742--1785,
De Gruyter Mouton, Berlin/Boston, 2011,
\ISBN 978-3-11-018523-2.

\reference Marr (1982):
David Marr,
\book{Vision: A Computational Inverstigation into
the Human Representation and Processing of Visual Information};
W.H.\ Freeman and Company, San Francisco, 1982.
\ISBN  978-0-7167-1284-8.

\reference McCarthy (1960):
John McCarthy,
``Recursive Functions of Symbolic Expressions
and Their Computation by Machine, Part I'';
in \periodical{Communications of the ACM},
vol.\ 3, no.\ 4, pp.\ 184--195, April 1960,
\DOI{10.1145/367177.367199}.

\reference McCarthy et al. (1962):
John McCarthy, Paul Abrahams, Daniel Edwards,
Timothy Hart, and Michael Levin,
 \book{LISP 1.5 Programmer's Manual};
The MIT Press,
Cambridge, Massachusetts, 1962,
\ISBN 978-0-262-13011-0.

\reference Miller (1956):
 George Miller,
``The Magical Number Seven, Plus or Minus Two:
Some Limits on Our Capacity for Processing Information'';
in \periodical{Psychological Review},
vol.\ 63, No.\ 2, pp.\ 81--97, 1956,
\DOI{10.1037/h0043158}.

\reference Pavlov (1927):
Ivan Pavlov,
\book{Conditioned Reflexes:
 An Investigation of the Physiological Activity
 of the Cerebral Cortex},
translated by G.V.\ Anrep;
Oxford University Press, London, 1927.
Reprinted by Dover, Minneola, NY, 1960, 2003;
\ISBN 0-486-43093-6.

\reference Pierce (1867):
Charles Sanders Peirce,
``On a New List of Categories'';
in \periodical{Proceedings of the American Academy
of Arts and Sciences},
vol.\ 7, pp.\ 287--298, 1868,
\DOI{10.2307/20179567}.
Presented May 14, 1867.

\reference Pinker \& Jackendoff (2004):
Steven Pinker, and Ray Jackendoff,
``The Faculty of Language: What's Special About It?'';
in \periodical{Cognition},
vol.\ 95, pp.\ 201--236, 2005,
\DOI{10.1016/j.cognition.2004.08.004}.
Received 16 January 2004; accepted 31 August 2004.

\reference Post (1936):
Emil Post,
``Finite Combinatory Processes --- Formulation 1'';
in \periodical{The Journal of Symbolic Logic},
vol.\ 1, no.\ 3, pp.\ 103--105, September 1936,
\DOI{10.2307/2269031}.
Received October 7, 1936.

\reference Post (1944):
Emil Post,
``Recursively Enumerable Sets of Positive Integers
  and their Decision Problems'';
in \periodical{Bulletin of the American Mathematical Society},
vol.\ 50, no.\ 5, pp.\ 284--316, 1944,
\DOI{10.1090/s0002-9904-1944-08111-1}.

\reference Rosser (1936):
J.\ Barkley Rosser,
``Extensions of Some Theorems of G\"odel and Church'';
in \periodical{Journal of Symbolic Logic},
vol.\ 1, pp.\ 87--91, 1936,
\DOI{10.2307/2269028}.

\reference Shieber (1986):
Stuart Shieber,  
``An Introduction to Unification-Based Approaches to Grammar'';
Microtome Publishing, Brookline, Massachusetts, 2003.
Reissue of the same title by
CSLI Publications, Stanford, California, 1986.\\
{\sc url:} \URL<http://nrs.harvard.edu/urn-3:HUL.InstRepos:11576719>.

\reference Simon \& Newell (1971):
Herbert Simon and Allen Newell,
``Human Problem Solving: The State of the Theory in 1970'';
in \periodical{American Psychologist},
 vol.\ 26, no.\ 2, pp.\ 145--159, February 1971,
\DOI{10.1037/h0030806}.

\reference Stabler (2014):
Edward Stabler,
``Recursion in Grammar and Performance'',\\
\DOIx{10.1007/978-3-319-05086-7\_8}%
{http://dx.doi.org/10.1007/978-3-319-05086-7_8};
in \book{Recursion: Complexity in Cognition},
Volume 43 of the series `Studies in Theoretical Psycholinguistics',
Tom Roeper, Margaret Speas (editors),
pp.\ 159--177,
Springer, Cham, Switzerland, 2014,
\ISBN 978-3-319-05085-0.

\reference Tomasello (2008):
Michael Tomasello,
\book{Origins of Human Communication};
The MIT Press, Cambridge, Massachusetts, 2008,
\ISBN 978-0-262-51520-7.

\reference Turing (1936):
Alan Turing,
``On Computable Numbers,
 with an Application to the Entscheidungsproblem'';
in \periodical{Proceedings of the London Mathematical Society},
vol.\ s2-42, no.\ 1, pp.\ 230--265, 1937,
\DOI{10.1112/plms/s2-42.1.230}.
Received 28 May, 1936. Read 12 November, 1936.

\reference Turing (1937):
Alan Turing,
``Computability and $\lambda$-Definability''; in
\periodical{The Journal of Symbolic Logic},
vol.\ 2, no.\ 4, pp.\ 153--163, December 1937,
\DOI{10.2307/2268280}.

\reference Vygotsky (1934):
Lev Vygotsky,
\book{Thought and Language},
newly revised and edited by Alex Kozulin;
The MIT Press, Cambridge, Massachusetts, 1986,
\ISBN 978-0-262-72010-6.

\reference Watumull et al. (2014):
Jeffrey Watumull, Marc Hauser, Ian Roberts \& Norbert Hornstein,
``On Recursion''; in
\periodical{Frontiers in Psychology},
vol.\ 4, article 1017, pp.\ 1--7, 2014,
\DOI{10.3389/fpsyg.2013.01017}.

\reference Zermelo (1908):
Ernst Zermelo,
``Untersuchungen \"uber die Grundlagen der Mengenlehre I'',
\DOIx{10.1007/978-3-540-79384-7\_6}%
{http://dx.doi.org/10.1007/978-3-540-79384-7_6};
in \book{Ernst Zermelo Collected Works} Volume I,
H.-D.\ Ebbinghaus, A.\ Kanamori (editors),
pp.\ 160--229,
Springer-Verlag, Berlin Heidelberg, 2010,
\ISBN 978-3-540-79383-0.

\bye